%% file: TemporalArgumentationFramework.tex
\documentclass[]{elsarticle}
\usepackage{lineno,hyperref}
\usepackage{xspace}
\usepackage{enumerate}
\usepackage{amssymb}
\usepackage{amsmath}
\usepackage{graphicx}
\usepackage{relsize}
\usepackage{color}
\usepackage{csquotes}
\usepackage{amssymb}
\usepackage{graphicx}
\usepackage{xspace}
\usepackage{dsfont}
\usepackage{textcomp}
\usepackage{url}
\usepackage{array}
\usepackage{pdflscape}
\usepackage{rotating}
\usepackage{blindtext}
\usepackage{longtable}
\usepackage{algpseudocode} 
\usepackage[ruled,vlined]{algorithm2e}
\usepackage{caption}
\usepackage{appendix}
\usepackage[all]{xy}
\usepackage{times}
\usepackage{float}

\input{Comandos}

\journal{Approximate Reasoning - https://doi.org/10.1016/j.ijar.2017.01.013}

\begin{document}

\begin{frontmatter}

    \title{Bipolar in Temporal Argumentation Framework}

    \author[label1,label2,label3]{Maximiliano C.\ D.\ Bud\'an}
    \author[label1]{Maria Laura Cobo}
    \author[label1,label2]{Diego C.\ Martinez}
    \author[label1]{Guillermo R.\ Simari}

    \address[label1]{Artificial Intelligence R\&D Laboratory $($LIDIA$)$, ICIC\\
             Universidad Nacional del Sur - Alem 1253, $($8000$)$ Bah\'{i}a Blanca, Buenos Aires}
    \address[label2]{Consejo Nacional de Investigaciones Cient\'ificas y T\'ecnicas\\
                     Av. Rivadavia 1917, Ciudad Aut\'onoma de Buenos Aires, Argentina}
    \address[label3]{Department of Mathematics - Universidad Nacional de Santiago del Estero\\
                     Belgrano(s) 1912, $($4200$)$ Capital, Sgo. del Estero, Argentina}

\begin{abstract}
	A \emph{Timed Argumentation Framework} (\emph{TAF}) is a formalism where arguments are only valid for consideration in a given period of time, called availability intervals, which are defined for every individual argument. The original proposal is based on a single, abstract notion of attack between arguments that remains static and permanent in time. Thus, in general, when identifying the set of acceptable arguments, the outcome associated with a \emph{TAF} will vary over time.
	
	In this work we introduce an extension of \emph{TAF} adding the capability of modeling a support relation between arguments. In this sense, the resulting framework provides a suitable model for different time-dependent issues. Thus, the main contribution here is to provide an enhanced framework for modeling a positive (support) and negative (attack) interaction varying over time, which are relevant in many real-world situations. This leads to a \emph{Timed Bipolar Argumentation Framework} (\emph{T-BAF}), where classical argument extensions can be defined. The proposal aims at advancing in the integration of temporal argumentation in different application domain.
\end{abstract}

\end{frontmatter}

\section{Introduction}\label{sec.intro}

Argumentation has contributed to the AI community with a human-like mechanism for the formalization of commonsense reasoning. Briefly speaking, argumentation can be associated with the interaction of arguments for and against a claim supported by some form of reasoning from a set of premises, with the purpose of ascertaining if that conclusion is acceptable~\cite{Bench-CaponD07,rahwan2009argumentation}. The study of this process suggested several argument-based formalisms dealing with applications in many areas such as legal reasoning, autonomous agents and multi-agent systems. In such environments, an agent may use argumentation to perform individual reasoning to reach a resolution over contradictory evidence or to decide between conflicting goals, while multiple agents may use dialectical argumentation to identify and settle differences interacting via diverse processes such as negotiation, persuasion, or joint deliberation. Many of such accounts of argumentation are based on Dung's foundational work characterizing \emph{Abstract Argumentation Frameworks} (\emph{AF})~\cite{Dung93,Dung95} where arguments are considered as atomic entities and their interaction is represented solely through an attack relation. 

In many cases, commonsense reasoning requires some representation of time since a notion of ``change" is relevant in the modeling of the argumentation capabilities of intelligent agents~\cite{AugustoS99,AugustoSimari01}. In particular, in~\cite{CoboMS11,CoboMartinezSimariNMR10,CoboMS10} a novel framework is proposed, called \emph{Timed Abstract Framework} (\emph{TAF}), combining arguments and temporal notions. In this formalism, arguments are relevant only in a period of time, called its availability interval. This framework maintains a high level of abstraction in an effort to capture intuitions related with the dynamic interplay of arguments as they become available and cease to be so. The notion of \textit{availability interval} refers to an interval of time in which the argument can be legally used for the particular purpose of an argumentation process. Thus, this kind of timed-argument has a limited influence in the system, given by the temporal context in which these arguments are taken into account. In \emph{TAF}, a skeptical timed interval-based semantics is proposed, using admissibility notions. As arguments may get attacked during a certain period of time, the notion of defense is also time-dependant, requiring a proper adaptation of classical acceptability. Furthermore, algorithms for the characterization of defenses between timed arguments are presented, used to specify the acceptability status of an argument varying over time~\cite{CoboMS10,CoboMartinezSimariNMR10}.

In most existing argumentation frameworks, only a conflict interaction between arguments is considered. However, in the last years, recent studies on argumentation have shown that a support interaction may exist between arguments, this kind of relation represents some real world situations. In this sense, several formal approaches were considered such as deductive support, necessary support and evidential support~\cite{cayrol2005acceptability,polberg2014,CohenGGS14,cayrol2015}, where a classical argumentative framework is enhanced to model a positive and negative interaction between arguments. In special, a simple abstract formalization of argument support is provided in the framework proposed by Cayrol and Lagasquie-Schiex in~\cite{cayrol2005acceptability}, called \emph{Bipolar Argumentation Framework} (\emph{BAF}), where they extend Dung's notion of acceptability by distinguishing two independent forms of interaction between arguments: support and attack. Besides the classical semantic consequences of attack, new semantic considerations are introduced that relies on the support of an attack and the attack of a support. 

In this work, we provide a timed argumentation framework to analyze the effect of attacks and supports in a dynamic discussion, leading place to a refined \emph{BAF}. In this sense, the resulting framework provides a suitable model for different time-dependent issues. The main contribution here is to provide an enhanced framework for modeling a positive (support) and negative (attack) interaction varying over time, which are relevant in many real-world situations. The aims of our work is to advance in the integration of temporal argumentation in different, time-related application domains and contribute to the successful integration of argumentation in different artificial intelligence applications, such as Knowledge Representation, Autonomous Agents in Decision Support Systems, and others of similar importance. Next, in order to state the relevance of our formalization, we analyse a classical example of bipolar argumentation case introduced in~\cite{amgoud2008bipolarity} about editorial publishing. Our formalism helps to represent a model that analyses the temporal effects, as follows:\\

\emph{Suppose a scenario where an Editorial is considering about presenting an important note related to a public person $\tt{P}$. For that, the chief editorial writer considers the following arguments, that are related to the importance and legality of the note.}

\begin{itemize}\itemsep 4pt
	\item[\argu{I}:] \emph{Information $\tt{I}$ concerning person $\tt{P}$ should be published.}
	\item[\argu{P}:] \emph{Information $\tt{I}$ is private so, $\tt{P}$ denies publication.} 
	\item[\argu{S}:] \emph{$\tt{I}$ is an important information concerning $\tt{P}$'s son.}
	\item[\argu{M}:] \emph{$\tt{P}$ is the new prime minister so, everything related to $\tt{P}$ is public.}
\end{itemize} 

\emph{Controversies arise during the above discussion. That is the case of the conflict between arguments \argu{P} and \argu{I}, and between arguments \argu{M} and \argu{P}. On the other hand, there is a relation between arguments \argu{P} and \argu{S}, which is clearly not a conflict. Moreover, \argu{S} provides a new piece of information enforcing argument \argu{P}.}\\

Although this is a proper example to introduce positive argument relations, it does not consider the evolution of time in an explicit way. Argumentation is a process by nature, and then it is interesting to evaluate arguments and conflicts in different stages of this process. In the previous example, from a temporal perspective, the analysis is made in a point of time where all arguments are available. We want to take into account the fact that arguments are introduced in different moments. Even more, some argument may become invalid or unusable over time. Thus, the editorial publishing example can be adapted in order to consider the evolution of information in time by making explicit the moments where those arguments can be used.\\

\emph{Based on the arguments presented previously, \argu{I} and \argu{P} can be both considered as general information applicable at any moment, a sort of editorial rules. However, the argument \argu{M} is available during the period of time where $\tt{P}$ is prime minister. Before that, argument \argu{M} does not apply. And after leaving the Primer Minister Office, the information about $\tt{P}$ may be less relevant for publication. Then, a new prime minister $\tt{P}_2$ may be a more important public person than $\tt{P}$, at least for media purposes.  The publisher may dismiss information about $\tt{P}$. Consider now that the chief editorial writer analyses a more complex scenario, taking some additional information into account in order to make a proper evaluation: the periods of time where $\tt{P}_2$ and $\tt{P}$ are prime ministers as well as the birth dates of their children, as follows:} 

\begin{itemize}\itemsep 4pt
	\item[\argu{I}:] \emph{Information $\tt{I}$ concerning person $\tt{P}$ should be published.}
	\item[\argu{P}:] \emph{Information $\tt{I}$ is private so, $\tt{P}$ denies publication.} 
	\item[\argu{S}:] \emph{$\tt{I}$ is an important information concerning $\tt{P}$'s son.}
	\item[\argu{T}:] \emph{$\tt{I}$ is an important information concerning $\tt{P}_2$'s son.}
	\item[\argu{M}:] \emph{$\tt{P}$ is the new prime minister so, everything related to $\tt{P}$ is public.}
	\item[\argu{N}:] \emph{$\tt{P}_2$ is the new prime minister so, everything related to $\tt{P}_2$ is public.}
\end{itemize}

\begin{center}
	$\begin{array}{|c|c|} 
	\hline
	Argument &  Temporal \ Availability\\ 
	\hline
	\hline
	\argu{I}    &  [0, \infty) \\
	\argu{P}  & [0, \infty)  \\ 
	\argu{S}  & [2013, Abr - 2013, Oct] \\ 
	\argu{T}   & [2012, Feb - 2012, Jun] \\ 
	\argu{M}  & [2012, Oct - 2014, Oct] \\ 
	\argu{N}   & [2010, Jun - 2012, Oct) \\  
	\hline
	\end{array} $
	\label{tab.temporalexample}
\end{center}

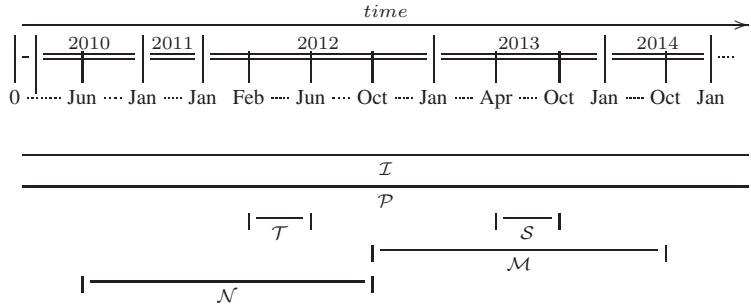
\begin{figure}[ht!]
	\begin{center}\leavevmode
		\xymatrix @R=0pc @C=0pc{
			&&&&&&&&&&&&&&&&&&&&&&&&&\\
			\ar@{->}[rrrrrrrrrrrrrrrrrrrrrrrr]^{time}&&&&&&&&&&&&&&&&&&&&&&&&\\
			&&&&&&&&&&&&&&&&&&&&&&&&\\
			\ar@{..}[r]&\ar@{=}^{2010}[rrrr]&&&&\ar@{=}^{2011}[rr]&&\ar@{=}^{2012}[rrrrrrr]&&&&&&&\ar@{=}^{2013}[rrrrr]&&&&&\ar@{=}^{2014}[rrr]&&&\ar@{..}[r]&&\\
			&&&&&&&&&&&&&&&&&&&&&&&&&\\
			\mbox{\footnotesize 0}\ar@{-}[uuuu]\ar@{..}[rrr]&\ar@{-}[uuuu]&&\mbox{\footnotesize Jun}\ar@{-}[uuu]\ar@{..}[rr]&&\mbox{\footnotesize Jan}\ar@{-}[uuuu]\ar@{..}[rr]&&\mbox{\footnotesize Jan}\ar@{-}[uuuu]&\mbox{\footnotesize Feb}\ar@{-}[uuu]\ar@{..}[rr]&&\mbox{\footnotesize Jun}\ar@{-}[uuu]\ar@{..}[rr]&&\mbox{\footnotesize Oct}\ar@{-}[uuu]\ar@{..}[rr]&&\mbox{\footnotesize Jan}\ar@{-}[uuuu]\ar@{..}[rr]&&\mbox{\footnotesize Apr}\ar@{-}[uuu]\ar@{..}[rr]&&\mbox{\footnotesize Oct}\ar@{-}[uuu]&\mbox{\footnotesize Jan}\ar@{-}[uuuu]\ar@{..}[rr]&&\mbox{\footnotesize Oct}\ar@{-}[uuu]&\mbox{\footnotesize Jan}\ar@{-}[uuuu]&&\\\\
			&&&&&&&&&&&&&&&&&&&&&&&&\\
			\ar@{-}[rrrrrrrrrrrrrrrrrrrrrrrr]_{\argu{I}}&&&&&&&&&&&&&&&&&&&&&&&&\\
			&&&&&&&&&&&&&&&&&&&&&&&&\\
			\ar@{-}[rrrrrrrrrrrrrrrrrrrrrrrr]_{\argu{P}}&&&&&&&&&&&&&&&&&&&&&&&&\\
			&&&&&&&&&&&&&&&&&&&&&&&&\\
			&&&&&&&&\ar@{-}[rr]_{\argu{T}}&&&&&&&&\ar@{-}[rr]_{\argu{S}}&&&&&&&&\\
			&&&&&&&&\ar@{-}[uu]&&\ar@{-}[uu]&&&&&&\ar@{-}[uu]&&\ar@{-}[uu]&&&&&&\\
			&&&&&&&&&&&&\ar@{-}[rrrrrrrrr]_{\argu{M}}&&&&&&&&&&&&\\
			&&&&&&&&&&&&\ar@{-}[uu]&&&&&&&&&\ar@{-}[uu]&&&\\
			&&&\ar@{-}[rrrrrrrrr]_{\argu{N}}&&&&&&&&&&&&&&&&&&&&&\\
			&&&\ar@{-}[uu]&&&&&&&&&\ar@{-}[uu]&&&&&&&&&&&&\\
		}
		\caption{Availability Distribution for the Arguments.}
		\label{GraphExample1}
	\end{center}
\end{figure}

\emph{The information about time is depicted in Figure \ref{GraphExample1}, where we show the time intervals in which every argument is \textit{available} or \textit{relevant} in a particular argumentative discussion.As we can see, the attack relation between arguments \argu{I} and \argu{P} is available in any moment of time, while the conflict between arguments \argu{M} and \argu{P} is active only in the time interval where \argu{M} is available, $[2012, Oct - 2014, Oct]$. On the other hand, argument \argu{M} reinforces argument \argu{I} in the time interval  $[2012, Oct - 2014, Oct]$, giving additional information about the person $\tt{P}$.}

In this example the time dimension is necessary to create a proper argumentation model that describes the evolution of the argumentative discussion. In this representation, we can analyse the different relationships between the arguments from a new perspective, such as: specification of time intervals where an argument is accepted, determination of moments in which an argument is strengthened by its supports (providing more information about a particular point) or to establish \textit{when} the supporting arguments provide extra conflict points for the supported argument. Furthermore, the proposed formalism allows the study of certain temporal properties associated with the arguments, such as their acceptability status over time.

This work is organized as follows: Section 2 presents a brief review of the classical bipolar abstract argumentation frameworks which allows the representation of support and conflict defined over arguments. In Section 3, we present some intuition to model time notions in the argumentative process. In Section 4, we introduce a concrete extension of the bipolar argumentation formalism where the temporal notion associated to the arguments is taken into account, and different temporal acceptability semantic process are presented. In Section 5, we present a real world example where the \emph{T-BAF}'s notions are applied in order to analyze a dynamic argumentation model. Finally, Section 6 and Section 7 are devoted to some related works, concluding remarks and further issues.

\section{Bipolar Abstract Argumentation}\label{sec.bipolar}

The basic idea behind argumentation is to construct arguments for and against a conclusion, analyse the general scenario, and then select the \textit{acceptable} ones. Arguments have different roles in front of each other. One might then say that arguments are presented in a ``bipolar" way since arguments in favour of a conclusion can be considered as positive while  arguments against the conclusion as negative ones. Based on these intuition, when representing the essential mechanism of argumentation the notion of bipolarity is a natural one. Abstracting away from the inner structure of the arguments, the Abstract Bipolar Argumentation Framework proposed by Cayrol and Lagasquie-Schiex in~\cite{cayrol2005acceptability}, extend Dung's notion of acceptability distinguishing two independent forms of interaction between arguments: support and attack. This new relation is assumed to be totally independent of the attack relation (i.e. it is not defined using the attack relation) providing a positive relation between arguments.

\begin{Definition}[Bipolar Argumentation Framework]\label{Def.Bipolar}	
	A Bipolar Argumentation Framework (\baf) is a 3-tuple $\Theta = \bipolar$, where \ard is a set of arguments, \atts  and \supp  are disjoint  binary relations on \ard called attack relation and support relation, respectively.
\end{Definition}

In order to represent a \baf, Cayrol and Lagasquie-Schiex extended the notion of graph presented by Dung in~\cite{Dung95} adding the representation of support between arguments. This argumentative model provides a starting point to analyse a  argumentative discussion enriched by the bipolarity of the human reasoning. This notion es defined as follows.

\begin{Definition}[Bipolar Argumentation Graph]\label{Def.BipolarGraph}
	Let $\Theta = \bipolar$ be a \baf. We define a directed graph for $\Theta$, denoted as \graphbipolar{\Theta}, taking as nodes the elements in \ard, and two types of arcs: one for  the attack relation (represented by plain arrows), and one for the support relation (represented by squid arrows).
\end{Definition}

In order to consider the interaction between supporting and defeating arguments, Cayrol and Lagasquie-Schiex in~\cite{cayrol2005acceptability} introduce the notions of \textit{supported} and \textit{secondary} defeat which combine a sequence of supports with a direct defeat. This notion is presented in the following definition.
\begin{Definition}[Supported and Secondary Defeat]~\label{def.defeat}
	Let $\Theta= \bipolar$ be an \baf, and $\argum{A}, \argum{B} \in \ard$ two arguments.
	
	\begin{itemize}
		\item[--] A supported defeat from \argum{A} to \argum{B} is a sequence $\argum{A}_1 \ \R_1 \ ... \ \R_n \ \argum{A}_n$, with $n \geq 3$, where $\argum{A}_1 = \argum{A}$ and $\argum{A}_n = \argum{B}$, such that $\forall i = 1 ... n-1$, $\R_i = \supp$ and $\R_{n-1} = \atts$.
		
		\item[--] A secondary defeat from \argum{A} to \argum{B} is a sequence $\argum{A}_1 \ \R_1 \ ... \ \R_n \ \argum{A}_n$, with $n \geq 3$, where $\argum{A}_1 = \argum{A}$ and $\argum{A}_n = \argum{B}$, such that $\R_{1} = \atts$ and $\forall i = 2 ... n-1$, $\R_i = \supp$.
	\end{itemize}
\end{Definition}
Note that, in \emph{BAF}, a sequence reduced to two arguments $\argum{A} \ \atts \ \argum{B}$ (a direct defeat $\argum{A} \to \argum{B}$) is also considered as a supported defeat from \argum{A} to \argum{B}.

\begin{Example}\label{ex.bipolarframework}
	Given a \baf $\Theta =\bipolar$, where:
	\begin{itemize}
		\item[] $\ard = \{\argum{A}; \argum{B}; \argum{C}; \argum{D}; \argum{E}; \argum{F}; \argum{G}; \argum{H}; \argum{I}; \argum{J} \}$,
		
		\item[] $\atts = \{(\argum{B},\argum{A}); (\argum{A},\argum{H});(\argum{C},\argum{B}); (\argum{G},\argum{I});(\argum{J},\argum{I});(\argum{F},\argum{C})\}$, and
		
		\item[] $\supp = \{(\argum{D},\argum{C}); (\argum{H},\argum{G});(\argum{I},\argum{F}); (\argum{E},\argum{B})\}$.
	\end{itemize}
	
	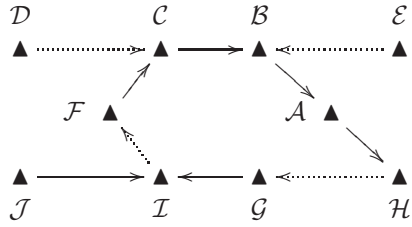
\begin{figure}[ht]
		\begin{center}\leavevmode
			\xymatrix @R=0pc @C=0pc{
				&{\argu{D}}&&&&&{\argu{C}}&&&&&{\argu{B}}&&&&&{\argu{E}} \\
				&{\blacktriangle}\ar@{..>}[rrrrr]&&&&&{\blacktriangle}\ar@{->}[rrrrr]&&&&&{\blacktriangle}\ar@{->}[rrddd]&&&&&{\blacktriangle}\ar@{..>}[lllll] \\
				&&&&&&&&&&&&&&&& \\
				&&&&&&&&&&&&&&&& \\
				&&&{\argu{F}}&{\blacktriangle}\ar@{->}[rruuu]&&&&&&&&{\argu{A}}&{\blacktriangle}\ar@{->}[rrrddd]&&& \\
				&&&&&&&&&&&&&&&& \\
				&&&&&&&&&&&&&&&& \\
				&{\blacktriangle}&&&&&{\blacktriangle}\ar@{<-}[lllll]\ar@{..>}[lluuu]&&&&&{\blacktriangle}\ar@{->}[lllll]&&&&&{\blacktriangle}\ar@{..>}[lllll] \\
				&{\argu{J}}&&&&&{\argu{I}}&&&&&{\argu{G}}&&&&&{\argu{H}} \\
			}
			
			\caption{Bipolar argumentation graph.}
			\label{Graph.Bipolar}
		\end{center}
		\vspace*{-15pt}
	\end{figure}
	
	We analyze the bipolar argumentation framework $\Theta$ characterized by the bipolar interaction graph depicted in \emph{Figure}~\ref{Graph.Bipolar}. For instance, \argum{J} and \argum{H} support defeat \argum{I}, since \argum{H} support G, and \argum{I} is attacked for \argum{G} and \argum{J} (direct attacker); in addition, \argum{J} and \argum{G} secondary defeat \argum{F}, because \argum{I} support \argum{F}, which is attacked for \argum{J} and \argum{G}. However, \argum{A} support defeat \argum{G} through \argum{H} support, and for that \argum{G} is defeated; also, \argum{B} support defeat \argum{A} (direct attacker), but \argum{D} support defeat \argum{B} through \argum{C}. Note that, the support defeat from \argum{G} to \argum{F} is invalidate since \argum{A} defeat \argum{H} which is a support of \argum{G}, irretrievably. 
\end{Example}

Cayrol and Lagasquie-Schiex in~\cite{cayrol2005acceptability} argued that a set of arguments must be in some sense coherent to model one side of an intelligent dispute. The coherence of a set of arguments is analyzed \emph{internally} (a set of arguments in which an argument attacks another in the same set is not acceptable), and \emph{externally} (a set of arguments which contains both a supporter and an attacker for the same argument is not acceptable). The internal coherence is captured extending the definition of \textit{conflict free set} proposed in~\cite{Dung95}, and  external coherence is captured with the notion of \textit{safe} set. 
\begin{Definition}[Conflict-free and Safe]\label{Def.ConflictSafe}
	Let $\Phi= \bipolar$ be an \baf, and $S \subseteq \ard$ be a set of arguments. 
	
	\begin{itemize}
		\item[--] $S$ is \emph{Conflict-free} iff $\nexists \argum{A}, \argum{B} \in S$ such that there is a supported or a secondary defeat from \argum{A} to \argum{B}.
		
		\item[--] $S$ is \emph{Safe} iff $\nexists \argum{A} \in \ard$ and  $\nexists \argum{B}, \argum{C} \in S$ such that there is a supported defeat or a secondary defeat from \argum{B} to \argum{A}, and either there is a sequence of support from \argum{C} to \argum{A}, or $\argum{A} \in S$.
	\end{itemize}
\end{Definition}
The notion of conflict-free in the above definition requires to take supported and secondary defeats into account, becoming a more restrictive definition than the classical version of conflict-freeness proposed by Dung. In addition, Cayrol and Lagasquie-Schiex show that the notion of safety is powerful enough to encompass the notion of conflict-freeness (\ie, if a set is safe, then it is also conflict-free). In addition, another requirement has been considered in~\cite{cayrol2005acceptability}, which concerns only the support relation, namely the closure under \supp.
\begin{Definition}[Closure in BAF] 
	Let $\Phi= \bipolar$ be an \baf. $S \subseteq \ard$ be a set of arguments. $S$ is closed under \supp iff $\forall \ \argum{A} \in S$, $\forall \ \argum{B} \in \ard$ if $\argum{A} \ \supp \ \argum{B}$ then $\argum{B} \in S$.
\end{Definition}
\begin{Example}[Continued Example~\ref{ex.bipolarframework}]~\label{ex.conflictsafe}
	The set $S_1 = \{\argum{I}; \argum{F}; \argum{D}; \argum{B}; \argum{K}\}$ is conflict free but not safe, since \argum{I} support defeat \argum{C} through \argum{F}, and \argum{D} support \argum{G}. The set $S_2 = \{\argum{J}; \argum{C}; \argum{D}; \argum{A}\}$ is conflict free and closed by \supp, then it is safe.
\end{Example}


Based on the previous concepts, Cayrol and Lagasquie-Schiex in~\cite{cayrol2005acceptability} extend the notions of defence for an argument with respect to a set of arguments, where they take into account the relations of support and conflict between arguments.
\begin{Definition}[Defence of $A$ from $B$ by $S$]
	Let $S \subseteq \ard$ be a set of arguments, and $\argum{A} \in \ard$ be an argument. $S$ defends collectively $A$ iff $\forall \ \argum{B} \in \ard$ if \argum{B} is a supported or secundary defeat of \argum{A} then $\exists \ \argum{C} \in S$ such that \argum{C} is a supported or secundary defeat of \argum{B}. In this case, it can be interpreted that \argum{C} defends \argum{A} from \argum{B}.
\end{Definition}
The authors proposed three different definitions for admissibility, from the most general to the most specific. The most general is based on Dung's admissibility definition; then, they extended  the notion of d-admissibility notion taking into account external coherence. Finally, external coherence is strengthened by requiring that admissible sets be closed for \supp.
\begin{Definition}[Admissibility in BAF]\label{Def.AdmissibilityBipolar}
	Let $\Phi = \bipolar$ be a \baf. Let $S \subseteq \ard$ be a set of arguments. The admissibility of a set $S$ is defined as follows:
	\begin{itemize}
		\item[--] $S$ is d-admissible if $S$ is conflict-free and defends all its elements.
		
		\item[--] $S$ is s-admissible if $S$ is safe and defends all its elements.
		
		\item[--] $S$ is c-admissible if $S$ conflict-free, closed for \supp and defends all its elements.
	\end{itemize}
\end{Definition}
\begin{Example}[Continued Example~\ref{ex.bipolarframework}]~\label{ex.admissible}
	The set $S_1 = \{\argum{J}; \argum{C};$ $\argum{D}; \argum{A}; \argum{E}\}$ is d-admissible, since it is conflict free and defend all its elements; however, it is not s-admissible, because \argum{C} and \argum{E} belong to $S_1$, where \argum{C} support defeat \argum{B} and \argum{E} support \argum{B}, and for that $S_1$ is not safe. It is important to note that, if a set of arguments not satisfies the s-admissibility, then not satisfies the c-admissibility; for that $S_1$ is not c-admissible. The set $S_2 = \{\argum{J}; \argum{C}; \argum{D}; \argum{A}\}$ is s-admissible, since it is safe and defend all its elements; in addition, it is closed for \supp, then $S_2$ is c-admissible too. 
\end{Example}

From the notions of coherence, admissibility, and extending the propositions introduced in~\cite{Dung95}, Cayrol and Lagasquie-Schiex in~\cite{cayrol2005acceptability} proposed different new semantics for the acceptability.
\begin{Definition}[Stable extension]\label{Def.StableBipolar}
	Let $\Phi = \bipolar$ be a \baf. Let $S \subseteq \ard$ be a set of arguments. $S$ is a {\em stable extension} of $\Phi$ if $S$ is conflict-free and for all $\argum{A} \notin S$, there is a supported or a secondary defeat of $\argum{A}$ in $S$.
\end{Definition}
\begin{Definition}[Preferred extension]\label{Def.PreferredBipolar}
	Let $\Phi = \bipolar$ be a \baf. Let $S \subseteq \ard$ be a set of arguments. $S$ is a d-preferred (resp. s-preferred, c-preferred) extension if $S$ is maximal (for set-inclusion) among the d-admissible (resp. s-admissible, c-admissible) subsets of \ard.
\end{Definition}
\begin{Example}[Continued Example~\ref{ex.bipolarframework}]~\label{ex.extensions}
	In our example, the set of arguments $S_1= \{\argum{J}; \argum{C}; \argum{D}; \argum{A}; \argum{E}\}$ is the stable extension, since there exist a defeater for the arguments \argum{I}, \argum{F}, \argum{G} and \argum{H} (as we explain in the \emph{Example~\ref{ex.bipolarframework}}). However, as we see in the \emph{Example~\ref{ex.admissible}}, this extension is not safe. 
	On the other hand, based on the \emph{Definition~\ref{Def.PreferredBipolar}}, we analyze the bipolar argumentation graph and determine the following preferred extensions: $S_1$ is a maximal d-admissible set, so $S_1$ is a d-preferred extension; $S_2 = \{\argum{J}; \argum{C}; \argum{D}; \argum{A}\}$ is a maximal s-admissible sets, so $S_2$ is a s-preferred extensions; and $S_2$ is a maximal c-admissible set, therefore $S_2$ is a c-preferred extension.
\end{Example} 

In the following section the support relation is considered among attacks in a timed context. Later, time-dependent semantics are presented.

\section{Towards a Temporal Argumentation Framework}

Our interest is to provide bipolar argumentation frameworks with a time-based notion of argument interaction. The focus is put in an abstract notion of \textit{availability} of arguments, which is a metaphor for a dynamic relative importance. Throughout this paper we mainly use the term ``available'', meaning that an argument will be considered just for a specific interval of time. 
However, availability can be interpreted in different ways. It may be the period of time in which an argument is relevant or strong enough or appropriate or any other suitable notion of \textit{relative importance} among arguments.  
The premise is that this availability is not persistent. In such a dynamic scenario, defeat and support may be sporadic and then proper time-based semantics need to be elaborated.

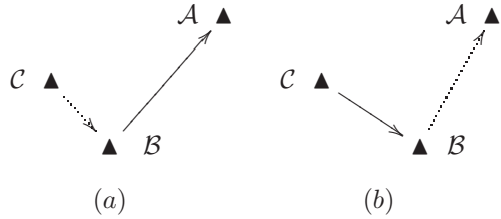
\begin{figure}[ht]
	\begin{center}\leavevmode
		\xymatrix @R=0pc @C=0pc{
			&&&&&&&& \\
			&&&&&&{\argu{A}}&{\blacktriangle}& \\
			&&&&&&&& \\
			&&&&&&&& \\
			&{\argu{C}}&{\blacktriangle}\ar@{..>}[dddrr]&&&&&& \\
			&&&&&&&& \\
			&&&&&&&& \\
			&&&&{\blacktriangle}\ar@{->}[rrruuuuuu]&{\argu{B}}&&& \\
			&&&&&&&& \\
			&&&&(a)&&&& \\
		}
		\xymatrix @R=0pc @C=0pc{
			&&&&&&&& \\
			&&&&&&{\argu{A}}&{\blacktriangle}& \\
			&&&&&&&& \\
			&&&&&&&& \\
			&{\argu{C}}&{\blacktriangle}\ar@{->}[dddrrr]&&&&&& \\
			&&&&&&&& \\
			&&&&&&&& \\
			&&&&&{\blacktriangle}\ar@{..>}[uuuuuurr]&{\argu{B}}&& \\
			&&&&&&&& \\
			&&&&(b)&&&& \\
		}
		\caption{Arguments Relations}
		\label{Graph.Relation}
	\end{center}
\end{figure}

Let \argu{A}, \argu{B} and \argu{C} be three arguments such that \argu{B} \atts \argu{A} and \argu{C} \supp \argu{B}, as shown in Figure \ref{Graph.Relation}(a). This is a minimal example of supported defeat. In the classical definition of bipolar argumentation framework, the set $S = \{\argu{C},\argu{B}\}$ is conflict-free. When considering availability of arguments, different conflict-free situations may arise. Suppose at moment $t_1$ arguments $\argu{C}$ and $\argu{A}$ are available while $\argu{B}$ is not. Then the set $S_1 = \{\argu{C},\argu{A}\}$ is conflict-free, since the attacker of $\argu{A}$ is not available \ie not relevant or strong at this particular moment. Suppose later at moment $t_2$ argument $\argu{B}$ becomes available. Then $S_1$ is no longer conflict-free since $\argu{C}$ supports a (now available) defeater of $\argu{A}$. Suppose later at moment $t_3$ argument $\argu{B}$ is not available again. Then set $S_1$ regains its conflict-free quality. Hence, a set of arguments in a timed context is not a conflict-free set by itself, but regarding certain moments in time. The set $S_1$ is conflict-free in $t_1$ and in $t_3$, and more generally speaking, in intervals of time in which availability of related arguments does not change.

In a similar fashion, consider the scenario of Figure \ref{Graph.Relation}(b), where \argu{B} \supp \argu{A} and \argu{C} \atts \argu{B}. Suppose at moment $t_1$ arguments $\argu{B}$ and $\argu{A}$ are available while $\argu{C}$ is not. Then the set $S_1 = \{\argu{A},\argu{B}\}$ is conflict-free. Suppose at moment $t_2$ arguments $\argu{C}$ is available, while $\argu{B}$ is not. Given the nature of the support relation, argument \argu{A} is not available too. Then, the set $S_1 = \{\argu{C}\}$ is conflict-free. Now, suppose that all the arguments are available at time $t_3$, then there is a conflict underlying in $\{\argu{C},\argu{A}\}$. 
In this case, argument $\argu{B}$ provides a support to $\argu{A}$, but in some moments of time $\argu{B}$ is attacked by the argument $\argu{C}$ providing a conflict for $\argu{A}$. This leads to the intuition that $\argu{B}$ is a weak support for $\argu{A}$ when $\argu{C}$ is available.

An interesting aspect of these situations is that, in a timed context, the concept of \textit{argument extension} must be revised.
Now the question is not \textit{whether} an argument is accepted (or rejected), but \textit{when}. 
Hence, a semantic analysis of the status of an argument must refer to intervals of time. 
We define a specific structure for this notion called \textit{t-profile}, to be presented later.

It is clear that in a dynamic environment, the set of conflict-free sets changes through time. Thus, the notion of acceptability in a bipolar argumentation scenario must be adapted when properly considered in a timed context. In this sense, we reformulate the attacks and supports notions defined in the classical bipolar argumentation framework, modeling the positive and negative effect of the arguments over time. In the following section the formal model of Timed Bipolar Argumentation Framework is introduced and the corresponding argument semantics are presented.

\section{Modeling Temporal Argumentation with T-BAF}\label{sec.taf}

The \textit{Timed Bipolar Argumentation Framework} (\tbaf) is an argumentation formalism where arguments are valid only during specific intervals of time, called \textit{availability intervals}. Attacks and supports are not permanent, since they are considered only when the involved arguments are \textit{available}. Thus, when identifying the set of acceptable arguments, the outcome associated with  a \tbaf may vary in time. 

In order to represent time, we assume that a correspondence was defined between the time line and the set of real numbers. A time interval, representing a period of time without interruptions, will be defined as follows 
For legibility reasons we use a different symbol as separator in the definition of intervals: ``$-$'' instead of, the tradicional, ``,''.
\begin{Definition}[Time Interval]\label{TimeInterval}
	A time interval $I$ represents a continuous period of time, identified by a pair of time-points. The initial time-point is called the startpoint of $I$, and the final time-point is called the endpoint of $I$. The intervals can be:
	
	\begin{itemize}
		\item[] Closed: defines a period of time that includes the definition points (startpoint and endpoint). Closed intervals are noted as $[a-b]$.
		
		\item[] Open: defines a period of time without the start and enpoint. Open intervals are noted as $(a-b)$.
		
		\item[] Semi-Closed: the periods of time includes one of the definition points but not both. Depending wich one is included, they are noted as $(a-b]$ (includes the endpoint) or $[a-b)$ (includes the startpoint).
	\end{itemize}
\end{Definition}
As it is usual, any of the previous intervals is considered empty if $b < a$, and the interval $[a-a]$ represents the point in time $\{a\}$. For the infinite endpoint, we use the symbol $+\infty$ and $-\infty$, as in $[a-{+\infty})$ or $({-\infty}- a]$ respectively, to indicate that there is no upper or lower bound for the interval respectively, and an interval containing this symbol will always be closed by $``)"$ or $``("$ respectively.

It will be necessary to group different intervals. 
We introduce the notion of \emph{time intervals set}. 

%
\begin{Definition}[Time Intervals Set]
	A time intervals set, or just intervals set, is a finite set $\mathcal{T}$ of time intervals.
\end{Definition}
Semantic elaborations are focused on the maximality of intervals.
For instance, two subsequent intervals may be joined and considered as one interval.  
When convenient, we will use the set of sets notation for time intervals sets. Concretely, a time interval set $\mathcal{T}$ will be denoted as the set of all disjoint and $\subseteq$-maximal individual intervals included in the set. For instance, we will use $\{(1-3],\ [4.5-8)\}$ to denote the time interval set $(1-3]\ \cup\ [4.5-8)$.

Now we formally introduce the notion of Timed Bipolar Argumentation Framework (\tbaf), which extends the \baf of Cayrol and Lagasquie-Schiex by incorporating an additional component, the availability function, which will be used to characterize those time intervals where arguments are available.
\begin{Definition}[Timed Bipolar Argumentation Framework]\label{def.TAF}
	A Timed Bipolar Argumentation framework (or simply \tbaf) is a triple $\Omega = \timebipolar$, where \ard is a set of arguments, \atts is a binary relation defined over \ard (representing attack), \supp is a binary relation defined over \ard (representing support), and $\av : \ard \longrightarrow \wp(\mathds{R})$ is an availability function for timed arguments, such that $\av(A)$ is the set of availability intervals of an argument $A$.
\end{Definition}
%
Note that since the arguments are only available during a certain period of time (the availability interval), it is rational to think that the relationship between arguments is relevant only when the arguments involved are available at the same time.

\begin{Definition}
	For any arguments \argum{A} and \argum{B}, we denote as \timedefeat{A}{B} and \timesupport{A}{B} the period of time in which the attack and the support between \argum{A} and \argum{B} is available, respectively.
\end{Definition}

\begin{Example}
	The corresponding timed bipolar argumentation framework of the introductory example is $\Omega_{intro}=\timebipolar$ where 
	\begin{itemize}
		\item \ard = $\{ {\cal I,P,S,T,M} \}$
		\item \atts = $\{ (\cal P,I) ,  (\cal M,P)\} $
		\item \supp = $\{ (\cal S,P)\} $
		\item $\av({\cal I})= [0,\infty]$, $\av({\cal P})= [0,\infty]$, $\av({\cal S})= [201304,201309]$,\\ 
		$\av({\cal T})= [201202,201206]$,  $\av({\cal M})= [201209,201409]$, \\ $\av({\cal S})= [201006,201209)$ 
	\end{itemize}
	Months and years are encoded to integers in order to preserve order.
\end{Example}

Some definitions are needed towards the formalization of the notion of \textit{acceptability} of arguments in \tbaf, which is  a time-based adaptation of the acceptability notions presented in Section 2 for \emph{BAF} with some new intuitions. 
First, we present the notion of \textit{t-profile}, binding an argument to a set of time intervals. 
This set represents intervals for special semantic consideration of the corresponding argument.
It is a structure that formalizes the phrase ``\textit{this argument, in those intervals}".
It is not necessarily the total availability of the argument as it is defined in the framework, so the reference has a special meaning when applied in appropriate contexts.
T-profiles constitute a fundamental component for the formalization of time-based acceptability, since it is the basic unit of timed reference for an argument.
\begin{Definition}[T-Profile]\label{def.T-profile}
	Let $\Omega = \timebipolar$ be a \tbaf. A timed argument profile for \argum{A} in $\Omega$, or just \tprotxt for \argum{A}, is a pair  $\tpro{A}$  where $\argum{A} \in \ard$ and $\tiempo{A}$ is a set of time intervals where \argum{A} is available, \ie, $\tiempo{A} \subseteq \av(\argum{A})$. The t-profile \tprobasic{A} is called the basic t-profile of \argum{A}.
\end{Definition}
The basic t-profile of an argument \argum{X} may be interpreted as the reference ``\argum{X}, whenever it is available".
Note that, as discussed previously, this argument may be attacked and defended as time evolves and then the basic t-profile will be probably fragmented under different semantics.

Since argument extensions are a collective construction based on arguments and interactions, several t-profiles will be considered. 
%
\begin{Definition}[Collection of T-Profiles]\label{def.Set-t-profile}
	Let $\Omega = \langle \ard, \atts,$ $ \supp,\av \rangle$ be a \tbaf. Let \tpron{X}{1}, \tpron{X}{2}, $\cdots$ , \tpron{X}{n} be \tprostxt. The set $C = \{$\tpron{X}{1},\tpron{X}{2}, $\cdots$ , \tpron{X}{n}$\}$ is a collection of \tprostxt iff it verifies the following conditions:
	
	\begin{itemize}
		\item[i\emph{)}] $\argum{X}_i \neq \argum{X}_j$ for all $i, j$ such that $i \neq j$, $1 \leq i,j \leq n$.
		
		\item[ii\emph{)}] $\tiempon{X}{i} \neq \emptyset$, for all $i$ such that $1 \leq i \leq n$.
	\end{itemize}
\end{Definition}

Given a collection of t-profiles, it will be sometimes necessary to reference all the arguments involved in those t-profiles.
\begin{Definition}[Arguments from a Collection of T-profiles]\label{def.cutprofiles}
	Let $C$ be a collection of t-profiles. The function \cutprofiles{Args}{C} defined as $\cutprofiles{Args}{C} = \{\argum{X} \mid \tpro{X} \in C\}$, obtain the set of arguments $Args$ involved in a collection of t-profiles $C$. 
\end{Definition}

Since arguments interact to each other, and every argument will be related to several intervals of time, it is necessary to introduce some basic operations. 
This is the intersection and inclusion of t-profiles, denoted as \textit{t-intersections} and \textit{t-inclusions}, formalized below:
\begin{Definition}[t-intersection]\label{def.tinterBudan}
	Let $\Omega = \timebipolar$ be a \tbaf. Let $C_1$ and $C_2$ be two collections of \tprostxt. We define the t-intersection of $C_1$ and $C_2$, denoted $C_1 \cap_t C_2$, as the collection of t-profiles  such that:\\
	
	\begin{small}
		\noindent$$C_1 \cap_t C_2 = \{(X, \tiempo{X} \cap \tiempoprima{X})\,|\, \tpro{X} \in C_1, \tproprima{X} \in C_2, \mathrm{and}\, \tiempo{X} \cap \tiempoprima{X} \neq \emptyset \}$$
	\end{small}
\end{Definition}
\begin{Definition}[t-inclusion]\label{def.tinclBudan}
	Let $C_1$ and $C_2$ be two collections of \tprostxt. We say that $C_1$ is \emph{t-included} in $C_2$, denoted as $C_1 \subseteq_t C_2$, if for any t-profile $\tpro{X} \in C_1$ there exists a t-profile $\tproprima{X} \in C_2$ such that $\tiempo{X} \subseteq \tiempoprima{X}$.
\end{Definition}  

In \tbaf, given a collection of t-profiles, it is possible generate a sequence of t-profiles from the existing relations between the arguments that are involved in them.
\begin{Definition}[Sequence of T-profiles]\label{def.secuenceprofiles}
	Let $\Omega = \langle \ard, \atts,$ $ \supp,\av \rangle$ be a \tbaf. Let $C = \{$ \tpron{X}{1}, \tpron{X}{2}, $\cdots$, \tpron{X}{n} $\}$ be a collection of t-profiles.
	Let $Args = \cutprofiles{Args}{C}$ be a set of arguments involved in the collection $C$. We will say that each t-profile of $C$ compose a secuence of t-profiles iff $\forall i = 1 \dots n-1$ verifies that $(\argum{A}_i,\argum{A}_{i+1}) \in \supp$ or $(\argum{A}_i,\argum{A}_{i+1}) \in \atts$, where $\argum{A}_i, \argum{A}_{i+1} \in Args$ and $\cap_t \tiempon{A}{i} \neq \emptyset \ \forall i = 1 \dots n$.
\end{Definition}

The following definitions reformulate \baf formalizations considering t-profiles instead of arguments. 
First, we will define the notion of supported and secundary defeat over time in \tbaf.
\begin{Definition}[Supported Defeat over Time]~\label{def.timesuppoerteddefeat}
	Let $\Omega= \timebipolar$ be a \tbaf. Let \tpro{A} and \tpro{B} two t-profiles. Let \tpron{A}{1} \tpron{A}{2} $\cdots$ \tpron{A}{n-1} \tpron{A}{n} be a sequence of t-profiles, with $n \geq 3$, $\tpron{A}{1} = \tpro{A}$ and \linebreak $\tpron{A}{n} = \tpro{B}$, such that $\forall i = 1 ... n-1$ $(A_i,A_{i+1}) \in \supp$ and $(A_{n-1},A_n) \in \atts$. The time interval in which \tpro{A} supported defeat \tpro{B}, denoted as \tsuppdefeat{A}{B}, is defined as $\tsuppdefeat{A}{B} = \cap^n_{i=1} \tiempon{A}{i}$.
\end{Definition}
A sequence reduced to two arguments $\argum{A} \ \atts \ \argum{B}$ (a direct defeat $\argum{A} \to \argum{B}$) is also considered as a supported defeat from $\argum{A}$ to $\argum{B}$.
\begin{Definition}[Secundary Defeat over Time]~\label{def.timesecundarydefeat}
	Let $\Omega= \timebipolar$ be a \tbaf. Let \tpro{A} and \tpro{B} two t-profiles. Let \tpron{A}{1} \tpron{A}{2} $\cdots$ \tpron{A}{n-1} \tpron{A}{n} be a sequence of t-profiles, with $n \geq 3$, $\tpron{A}{1} = \tpro{A}$ and \linebreak $\tpron{A}{n} = \tpro{B}$, such that $(A_1,A_2) \in \atts$ and $\forall i = 2 ... n$, $(A_i,A_{i+1}) \in \supp$. We will define the time interval in which \tpro{A} secundary defeat \tpro{B}, denoted as \tsecdefeat{A}{B}, is defined as $\tsecdefeat{A}{B} = \cap^n_{i=1} \tiempon{A}{i}$.
\end{Definition}

\begin{Example}
	We will introduce an abstract example, through which we will clarify the concepts introduced until now. In this case we introduce the notion of time availability into the arguments presented in Example~\ref{ex.bipolarframework}.\\
	
	\noindent Given a \tbaf $\Omega =\timebipolar$, where:
	
	\begin{itemize}
		\item[] $\ard = \{\argum{A}; \argum{B}; \argum{C}; \argum{D}; \argum{E}; \argum{F}; \argum{G}; \argum{H}; \argum{I}; \argum{J} \}$,
		
		\item[] $\atts = \{(\argum{B},\argum{A}); (\argum{A},\argum{H}); (\argum{C},\argum{B}); (\argum{G},\argum{I});(\argum{J},\argum{I});(\argum{F},\argum{C})\}$,
		
		\item[] $\supp = \{(\argum{D},\argum{C}); (\argum{H},\argum{G});(\argum{I},\argum{F}); (\argum{E},\argum{B})\}$, and
		
		\item[] $\av = \{ \tproins{A}{[0-100]}; \tproins{B}{(90-150]}; \tproins{C}{[30-180]}; \tproins{D}{[0-60]};$\linebreak 
		$\tproins{E}{[100-160)}; \tproins{F}{[50-90]}; \tproins{G}{[60-120]}; \tproins{H}{[40-80]};$ \linebreak
		$\tproins{I}{(70-110]}; \tproins{J}{[0-90)}\}$.
	\end{itemize}
	
	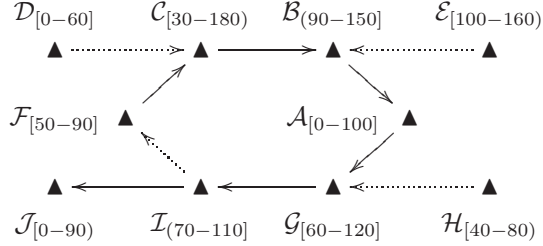
\begin{figure}[ht]
		\begin{center}\leavevmode
			\xymatrix @R=0pc @C=0pc{
				&{\argu{D}}_{\left[0-60\right]}&&{\argu{C}}_{\left[30-180\right)}&&{\argu{B}}_{\left(90-150\right]}&&{\argu{E}}_{\left[100-160\right)} \\
				&{\blacktriangle}\ar@{..>}[rr]&&{\blacktriangle}\ar@{->}[rr]&&{\blacktriangle}\ar@{->}[rddd]&&{\blacktriangle}\ar@{..>}[ll] \\
				&&&&&&& \\
				&&&&&&& \\
				&{\argu{F}}_{\left[50-90\right]}&{\blacktriangle}\ar@{->}[ruuu]&&&{\argu{A}}_{\left[0-100\right]}&{\blacktriangle}\ar@{->}[lddd]& \\
				&&&&&&& \\
				&&&&&&& \\
				&{\blacktriangle}&&{\blacktriangle}\ar@{->}[ll]\ar@{..>}[luuu]&&{\blacktriangle}\ar@{->}[ll]&&{\blacktriangle}\ar@{..>}[ll] \\
				&{\argu{J}}_{\left[0-90\right)}&&{\argu{I}}_{\left(70-110\right]}&&{\argu{G}}_{\left[60-120\right]}&&{\argu{H}}_{\left[40-80\right)} \\
			}
			\caption{Bipolar argumentation graph with Timed Availability}
			\label{Graph.TimeBipolar}
		\end{center}
		\vspace*{-15pt}
	\end{figure}
	
	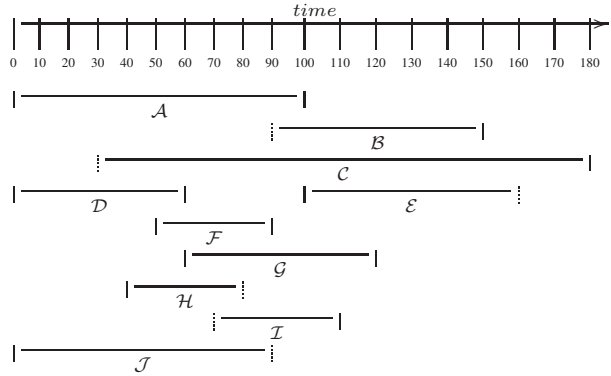
\begin{figure}[ht]
		\begin{center}\leavevmode
			\xymatrix @R=0pc @C=0pc{
				&&&&&&&&&&&&&&&&&&&\\
				\ar@{->}[rrrrrrrrrrrrrrrrrrr]^{time}&&&&&&&&&&&&&&&&&&&\\
				&&&&&&&&&&&&&&&&&&&\\
				\mbox{\tiny 0}\ar@{-}[uuu]&\mbox{\tiny 10}\ar@{-}[uuu]&\mbox{\tiny 20}\ar@{-}[uuu]&\mbox{\tiny 30}\ar@{-}[uuu]&\mbox{\tiny 40}\ar@{-}[uuu]&\mbox{\tiny 50}\ar@{-}[uuu]&\mbox{\tiny 60}\ar@{-}[uuu]&\mbox{\tiny 70}\ar@{-}[uuu]&\mbox{\tiny 80}\ar@{-}[uuu]&\mbox{\tiny 90}\ar@{-}[uuu]&\mbox{\tiny 100}\ar@{-}[uuu]&\mbox{\tiny 110}\ar@{-}[uuu]&\mbox{\tiny 120}\ar@{-}[uuu]&\mbox{\tiny 130}\ar@{-}[uuu]&\mbox{\tiny 140}\ar@{-}[uuu]&\mbox{\tiny 150}\ar@{-}[uuu]&\mbox{\tiny 160}\ar@{-}[uuu]&\mbox{\tiny 170}\ar@{-}[uuu]&\mbox{\tiny 180}\ar@{-}[uuu]\\
				&&&&&&&&&&&&&&&&&&&&\\
				\ar@{-}[rrrrrrrrrr]_{\argu{A}}&&&&&&&&&&&&&&&&&&&&\\
				\ar@{-}[uu]&&&&&&&&&&\ar@{-}[uu]&&&&&&&&&\\
				&&&&&&&&&\ar@{-}[rrrrrr]_{\argu{B}}&&&&&&&&&&\\
				&&&&&&&&&\ar@{..}[uu]&&&&&&\ar@{-}[uu]&&&&\\
				&&&\ar@{-}[rrrrrrrrrrrrrrr]_{\argu{C}}&&&&&&&&&&&&&&&&\\
				&&&\ar@{..}[uu]&&&&&&&&&&&&&&&\ar@{-}[uu]&\\
				\ar@{-}[rrrrrr]_{\argu{D}}&&&&&&&&&&\ar@{-}[rrrrrr]_{\argu{E}}&&&&&&&&&\\
				\ar@{-}[uu]&&&&&&\ar@{-}[uu]&&&&\ar@{-}[uu]&&&&&&\ar@{..}[uu]&&&\\
				&&&&&\ar@{-}[rrrr]_{\argu{F}}&&&&&&&&&&&&&&\\
				&&&&&\ar@{-}[uu]&&&&\ar@{-}[uu]&&&&&&&&&&\\
				&&&&&&\ar@{-}[rrrrrr]_{\argu{G}}&&&&&&&&&&&&&\\
				&&&&&&\ar@{-}[uu]&&&&&&\ar@{-}[uu]&&&&&&&\\
				&&&&\ar@{-}[rrrr]_{\argu{H}}&&&&&&&&&&&&&&&\\
				&&&&\ar@{-}[uu]&&&&\ar@{..}[uu]&&&&&&&&&&&\\
				&&&&&&&\ar@{-}[rrrr]_{\argu{I}}&&&&&&&&&&&&\\
				&&&&&&&\ar@{..}[uu]&&&&\ar@{-}[uu]&&&&&&&&\\
				\ar@{-}[rrrrrrrrr]_{\argu{J}}&&&&&&&&&&&&&&&&&&&\\
				\ar@{-}[uu]&&&&&&&&&\ar@{..}[uu]&&&&&&&&&&\\
			}
			\caption{Temporal Distribution}
			\label{Graph.DistTime}
		\end{center}
		\vspace*{-15pt}
	\end{figure}
	
	Next, we analyze the timed bipolar argumentation framework $\Omega$ characterized by the bipolar interaction graph depicted in \emph{Figure~\ref{Graph.TimeBipolar}}, and temporal distribution depicted in \emph{Figure~\ref{Graph.DistTime}}. In particular, we will pay attention to the relations that arise from the leaf nodes of the graph, in order to clarify the \emph{Definitions}~\ref{def.timesuppoerteddefeat} and~\ref{def.timesecundarydefeat}.
	On one hand, \argum{J} support defeat \argum{I} in the time intervals $\tsuppdefeat{J}{I} = \tiempo{J} \cap \tiempo{I} = \{(70-90)\}$, and \argum{J} secondary defeat \argum{F} in the time intervals $\tsuppdefeat{J}{F} = \tiempo{J} \cap \tiempo{I} \cap \tiempo{F} = \{(70-90)\}.$ Also, analysing the leaf argument \argum{D}, the support defeat from  \argum{D} to \argum{B} is invalidate since the time interval $\tsuppdefeat{D}{B} = \{\emptyset\}$, where $\tsuppdefeat{D}{B} = \tiempo{A} \cap \tiempo{C} \cap \tiempo{B} = \{\emptyset\}.$ On the other hand, from the argument \argum{E}, there exist a support defeat from \argum{E} to \argum{A} in the time interval $\tsuppdefeat{E}{A} = \tiempo{E} \cap \tiempo{B} \cap \tiempo{A} = \{[100-100]\}.$
	
\end{Example}

Once defined the relations of attack over time using the t-profiles, we are able to adapt the notions of conflict-free and safeness used in \baf, now considering time.
\begin{Definition}[Conflict-free and Safe]\label{Def.ConflictSafetime}
	Let $\Omega= \langle \ard,$ $\atts,$ $ \supp,\av \rangle$ be a \tbaf, and $S$ be a collection of t-profiles defined for $\Omega$.
	
	\begin{itemize}
		\item[--] $S$ is \emph{Conflict-free} iff $\nexists \tpro{A}, \tpro{B} \in S$ such that $\tsuppdefeat{A}{B} \neq \emptyset$ or $\tsecdefeat{A}{B} \neq \emptyset$.
		
		\item[--] $S$ is \emph{Safe} iff $\nexists \tpro{A}, \tpro{B} \in S$ and  $\nexists \tpro{C}$ where $\tpro{C}$ is a valid $\Omega$'s t-profile such that $\tsuppdefeat{A}{C} \neq \emptyset$ or $\tsecdefeat{A}{C} \neq \emptyset$, and either there is a sequence of support from \tpro{B} to \tpro{C}, or $\tpro{C} \in S$.
	\end{itemize}
\end{Definition}
In addition, another requirement has been considered in~\cite{cayrol2005acceptability}, which concerns only the support relation, namely the \textit{closure under} \supp. 
This is presented in a timed context as follows.
\begin{Definition}[Closure in T-BAF] 
	Let $\Omega= \langle \ard,$ $\atts,$ $ \supp,\av \rangle$ be a \tbaf, and $S$ be a collection of t-profiles defined for $\Omega$. The set $S$ is closed under \supp iff $\forall \ \tpro{A} \in S$, $\forall \ \tpro{B}$ where $\tpro{B}$ is a valid $\Omega$'s t-profile: if $\tpro{A} \ \supp \ \tpro{B}$ where $\tiempo{A} \cap \tiempo{B} \neq \emptyset$  then at least $ \langle\argum{B}, \tiempo{A} \cap \tiempo{B}\rangle \in S$.
\end{Definition}
\begin{Example} 
	The collection $C_1 = \{\tproins{A}{[0-100]} ; \tproins{C}{[30-50) , (90-180)} ; \linebreak \tproins{D}{[0-60]} ; \tproins{E}{[100-160)} ; \tproins{F}{[50-90]} ;$ $\tproins{G}{[80-120]} ; \linebreak \tproins{J}{[0-90)}\}$ is conflict-free but not safe, since the argument \argum{D} support \argum{C} and \argum{F} attacks \argum{C} in the time interval set $\{[50-60]\}$; On another case, the argument \argum{B} is supported and attacked by \argum{E} and \argum{C}, respectively, in the time interval set $\{[100-150]\}$. The collection $C_2 = \{\tproins{A}{[0-100]} ; \tproins{D}{[0-50)} ;  \tproins{C}{[30-50),(90-100), \linebreak (150-180)}; \tproins{E}{(150-160)} ;\tproins{F}{(60-90]} ; \tproins{G}{[80-120]} ; \tproins{J}{[0-90)}\}$ is conflict-free and safe.
\end{Example}

\begin{Proposition}
	Let $S$ be a collection of t-profiles:
	\begin{itemize} 
		\item[--] If $S$ is safe, then any collection $S' \subseteq_t S$ is conflict-free. 
		\item[--] If $S$ is conflict-free and closed for \supp then $S$ is safe.
	\end{itemize}
\end{Proposition}

Following the same fashion, the following definitions reformulate \baf notions considering t-profiles instead of arguments. 
We define the \textit{defense} of an argument over time, taking into account the corresponding support and secondary defeat.
\begin{Definition}[Defense of \argum{A} from \argum{B} by a collection $C$]
	Let $\Omega= \timebipolar$ be a \tbaf, and $C$ be a conflict-free collection of t-profiles. Let \tpro{A} and \tpro{B} two t-profiles, where \argum{B} attacks \argum{A} through a support or secondary attacks such that $\tsecdefeat{B}{A} \neq \emptyset$ and/or $\tsuppdefeat{B}{A} \neq \emptyset$. The defense t-profile of \argum{A} from \argum{B} with respect to $C$, denoted as \tdefence{A}{B} is defined as follows:
	
	\begin{center}
		$\tdefence{A}{B} =_{\mathit{def}} \tiempobasic{A} \cap \left( \tsupdefence{A}{B} \cup \tsecdefence{A}{B}\right) $ 
	\end{center}
	
	\noindent where $\tsupdefence{A}{B} =_{\mathit{def}} \bigcup_{\tiny \argum{C}\in{\{\argum{X}\,|\,\tpro{X} \in C,\,(\argum{X},\tsuppdefeat{X}{B} \neq \emptyset\}}} \tsuppdefeat{C}{B} $\\ and
	$\tsecdefence{A}{B} =_{\mathit{def}} \bigcup_{\tiny \argum{C}\in{\{\argum{X}\,|\,\tpro{X} \in C,\,(\argum{X},\tsecdefeat{X}{B}\neq \emptyset\}}} \tsecdefeat{C}{B}$.
\end{Definition}

Intuitively, \argum{A} is defended from the attack of \argum{B} when: (a) \argum{B} is not available and (b) in those intervals where the attacker \argum{B} is available but it is in turn attacked by an available argument \argum{C} in the collection $C$. 

\begin{Definition}[Acceptable t-profile of \argum{A} w.r.t. $C$] Let $\Omega= \timebipolar$ be an \tbaf. The acceptable t-profile for \argum{A} w.r.t. $C$, denoted as \tdefences{A}, is defined as follows:
	\begin{center}
		$\tdefences{A} =_{\mathit{def}}\bigcap_{\tiny \argum{B}\in{\{\argum{X}\,|\,\tsuppdefeat{X}{A} \neq \emptyset \ \vee \ \tsecdefeat{X}{A} \neq \emptyset\}}}$\\ $ \left( \tiempobasic{A} \setminus \left( \tsuppdefeat{B}{A} \cup \tsecdefeat{B}{A}\right) \right)  \cup \tdefence{A}{B} $
	\end{center}
	where  \tdefence{A}{B} is the time interval where \argum{A} is defended of its attacker \argum{B} by $C$. Then, the intersection of all time intervals in which \argum{A} is defended from each of its attackers by the collection $C$, is the time interval where \argum{A} is available and is acceptable with respect to $C$.
\end{Definition}
\begin{Example}
	In this example, we will show how the acceptable t-profile of \argum{I} from a collection $C_3 = \{\tproins{A}{[0-100]}$ is calculated. 
	
	$\tdefencesBis{I}{C_3} = \left( \tiempobasic{I} \setminus\left( \tsecdefeat{G}{I} \cup \tsecdefeat{H}{I}\right) \right)  \cup \left( \tdefenceBis{A}{G}{C_3} \cup \tdefenceBis{A}{H}{C_3}\right)  = \\
	= (70-110] \setminus ( (70-110] \cup (70-80))) \cup  ( (70-110] \cup (70-80)]) = (70-110]$ 
\end{Example}

In this section, we adapt the three different definitions for admissibility proposed by Cayrol and Lagasquie-Schiex in~\cite{cayrol2005acceptability}, through the new version of conflict-freeness and safety.
\begin{Definition}[Admissibility in T-BAF]\label{Def.admissibilityovertime}
	Let $\Omega= \langle \ard, \atts, $ $\supp,\av \rangle$ be a \tbaf. Let $C$ be a collection of t-profiles. The admissibility of a collection $C$ is defined as follows:
	
	\begin{itemize}\small\itemsep 4pt
		\item[--] $C$ is td-admissible if $C$ is conflict-free and defends all its elements.
		
		\item[--] $C$ is ts-admissible if $C$ is safe and defends all its elements.
		
		\item[--] $C$ is tc-admissible if $C$ conflict-free, closed for \supp and defends all its elements.
	\end{itemize}
\end{Definition}
\begin{Example}	
	The collection $C_4 = \{\tproins{A}{[0-100)} ; \tproins{C}{[30-50) , (70-180)} ;$\linebreak $\tproins{D}{[0-60]} ; \tproins{E}{[100-160)} ; \tproins{F}{[50-70]} ;$ $\tproins{G}{(80-120]} ;$\linebreak $\tproins{J}{[0-90)}\}$ is td-admissible since it is conflict-free and defends all its elements over time. However, $C_4$ is not a ts-admissible or tc-admissible collection of t-profiles.  $C_5 = \{\tproins{A}{[0-100]} ; \tproins{C}{[30-50) , (70-180)} ; \tproins{D}{[0-60]} ;$ \linebreak$\tproins{E}{(150-160)} ; \tproins{F}{[50-70]} ;\tproins{G}{(80-120]} ; \tproins{J}{[0-90)}\}$ is ts-admissible because its safe and defends all its elements. In addition, $C_5$ is closed for \supp, then it is tc-admissible.
\end{Example}

\begin{Proposition}
	Let $\Omega= \langle \ard,$ $ \atts, \supp,\av \rangle$ be a \tbaf, then:
	\begin{itemize} 
		\item[--] A td-admissible extension is t-included in a ts-admissible extension. 
		\item[--] A ts-admissible extension is t-included in a tc-admissible extension.
	\end{itemize} 
\end{Proposition}

Now we can define the classic argument semantics for \tbaf. 
First, we present the stable extension with a dynamic intuition. 
Then we introduce an adapted, timed version of preferred extension. 
Each one has a special property based on the admissibility notion.
\begin{Definition}[Stable extension over Time]\label{Def.StableBipolartime}
	Let $\Omega= \langle \ard,$ $ \atts, \supp,\av \rangle$ be a \tbaf. Let $C$ be a collection of t-profiles. $C$ is a {\em t-stable extension} of $\Omega$ if $C$ is conflict-free and for all $\tpro{A} \notin C$, verifies that $\tiempo{A} \ \setminus \ (\bigcup \tsecdefeat{B}{A} \ \cup \ \bigcup \tsuppdefeat{B}{A}) = \emptyset$ for all $\tpro{B} \in C$.
\end{Definition}
\begin{Definition}[Preferred extension over Time]\label{Def.PreferredBipolartime}
	Let $\Omega= \timebipolar$ be an \tbaf. Let $C$ be a collection of t-profiles. $C$ is a td-preferred (resp. ts-preferred, tc-preferred) extension if $C$ is maximal (for set-t-inclusion) among the td-admissible (resp. ts-admissible, tc-admissible).
\end{Definition}

The relations between t-preferred extensions and t-stable extensions are stated in the following proposition. 
Note that these are consistent with the classical \baf.
%
\begin{Proposition}
	Let $\Omega= \langle \ard,$ $ \atts, \supp,\av \rangle$ be a \tbaf, then:
	
	\begin{itemize}
		\item[-] A td-preferred extension is t-included in a ts-preferred extension.
		
		\item[-] A ts-preferred extension is t-included in a tc-preferred extension.
		
		\item[-] A ts-preferred extension closed under \supp is also a tc-preferred.
		
		\item[-] A td-preferred extension is t-included in a t-stable extension.
		
		\item[-] A ts-preferred extension is t-included in a safe t-stable extension.
		
		\item[-] A tc-preferred extension is t-included in a safe t-stable extension.
	\end{itemize}
\end{Proposition}

Given an \tbaf $\Omega= \timebipolar$, and an argument $\argum{A} \in \ard$, we will use $t\mbox{-}PR_d(\argum{A})$, $t\mbox{-}PR_s(\argum{A})$, $t\mbox{-}PR_c(\argum{A})$ and $t\mbox{-}ES(\argum{A})$ to denote the set of intervals on which \argum{A} is acceptable in $\Omega$ according to td-preferred, ts-preferred, tc-preferred and t-stable semantics respectively, using again the skeptical approach where it corresponds. The following property establishes a connection between acceptability in our extended temporal framework T-BAF and acceptability in Cayrol and Lagasquie-Schiex’s frameworks. 
\begin{Theorem}
	Let $\Omega= \timebipolar$ be a \tbaf and let $\alpha$ representing a point in time. Let $\Theta'_{\alpha} = \langle \ard'_{\alpha}, \atts^{\alpha}, \supp^{\alpha} \rangle$ be a
	bipolar abstract framework obtained from $\Omega$ in the following way: $\ard'_{\alpha} = \{\argum{A} \in \ard \mid \alpha \in \tiempo{A}\}$,  $\atts^{\alpha} = \{(\argum{A}, \argum{B})\in \alpha \in \timedefeat{A}{B}\}$ and $\supp^{\alpha} = \{(\argum{A}, \argum{B})\in \alpha \in \timesupport{A}{B}\}$.
	Let $C$ a collection of t-profiles in $\Omega$, and $C'_{\alpha} = \{\argum{A} \mid  \tdefencese{A} \in C$ and $\alpha \in \tdefencese{A}\}$. It holds that, if $C$ is an td-preferred extension (resp. ts-preferred, tc-preferred, and t-stable) w.r.t. $\Omega$, then $C'_{\alpha}$ is a  d-preferred extension (resp. ts-preferred, tc-preferred, and t-stable) w.r.t. $\Theta'_{\alpha}$.
\end{Theorem} 
Intuitively, the BAF $\Theta'_{\alpha}$ represents a snapshot of the T-BAF framework $\Omega$ at the time point $\alpha$, where the arguments and attacks in $\Theta'_{\alpha}$ are those that are available at the time point $\alpha$ in $\Omega$. Then, this Lemma states that an td-preferred extension (resp. ts-preferred, tc-preferred, and t-stable) $C$ for T-BAF at the time point $\alpha$ coincides with a d-preferred extension $C'_{\alpha}$ (resp. ts-preferred, tc-preferred, and t-stable) of $\Theta'_{\alpha}$.\\

In addition, we formally establish that two arguments with an attack path cannot coincide in time when both belong to the same extension in a given semantics.
\begin{Proposition}
	Let $\Omega= \langle \ard,$ $ \atts, \supp,\av \rangle$ be a \tbaf, and \tpro{A} and \tpro{B} be two t-profiles, where \argum{B} defeats \argum{A} through a support or secondary attacks, then it holds that:
	
	\begin{itemize}
		\item[--] $t\mbox{-}ES(\argum{A}) \cap t\mbox{-}ES(\argum{B}) = \emptyset$
		
		\item[--] $t\mbox{-}PR_d(\argum{A}) \cap t\mbox{-}PR_d(\argum{B}) = \emptyset$;
		
		\item[--] $t\mbox{-}PR_s(\argum{A}) \cap t\mbox{-}PR_s(\argum{B}) = \emptyset$;
		
		\item[--] $t\mbox{-}PR_c(\argum{A}) \cap t\mbox{-}PR_c(\argum{B}) = \emptyset$; and
	\end{itemize}
\end{Proposition}

\begin{Example}
	In our example, the set of arguments $C_4$ is the stable extension, since there exist a defeater for each t-profile that does not belong to $C_4$. In addition, $C_4$ is a td-preferred extension since it is the maximal td-admissible collection of t-profiles that defends all of its elements. On another hand, $C_5$ is a ts-preferred extension because it is the maximal ts-admissible collection of t-profiles that defends all of its elements. Also, $C_5$ is closed by \supp, then it is tc-preferred extension.
\end{Example} 

It is worthwhile to notice that when time becomes irrelevant, \ie\ reduced to a particular instant or all arguments are available in exactly the same periods of time, the behavior of $\tbaf$ is equivalent to the original $\baf$. 

\section{Application Example}\label{sec.exampletaf}

As stated before, the aim of this work is to increase the representational capability of \emph{BAF}s, by adding a temporal dimension towards a model of dynamic argumentation discussion. 
To illustrate the usefulness of this direction in the context of agent and multi-agents systems, we discuss an example where the formalism provides a better characterization of the overall situation.\smallskip

\emph{Consider the following scenario where an agent is looking for an apartment to rent. As expected, while considering a  candidate she analyzes different arguments for and against renting it. 
	These arguments are subject to availability or relevance in time. 
	The task is to determine in the present (time 0) if the property is a good option in the future, counting with 150 days to make such a decision. 
	The arguments and the availability intervals follows.}
\begin{itemize}\itemsep 0pt
	
	\item[\argum{A}] \textit{She should rent it, the apartment has a good location since it is near of her work.}$\small{[0-150]}$
	
	\item[\argum{B}] \textit{The apartment is located in an well illuminated and safe area.}$\small{[0-150]}$
	
	\item[\argum{C}] \textit{The property is in a quiet area, because most of the neighbors are retirees and peaceful people.}$\small{[0-150]}$
	
	\item[\argum{D}] \textit{The apartment is small; therefore, she should not rent it.}$\small{[0-150]}$
	
	\item[\argum{E}] \textit{Despite the apartment size, the spaces are well distributed.}$\small{[0-150]}$
	
	\item[\argum{F}] \textit{She should not rent it, since the apartment seems to have humidity problems.}$\small{[0-150]}$
	
	\item[\argum{G}] \textit{There are rumours that a nightclub will open in the area in 50 days, so the area will not be quiet anymore.}$\small{[50-150]}$
	
	\item[\argum{H}] \textit{The humidity problems are difficult and costly to resolve.}$\small{[0-150]}$
	
	\item[\argum{I}] \textit{Laws forbid the opening of a nightclub in this urban area, but this will be revised in the next Town Hall meeting.}$\small{[0-80]}$
	
	\item[\argum{J}] \textit{The person responsible for maintenance is committed to fixing the humidity problem at a low cost.}$\small{[0-150]}$
\end{itemize}

Next, we instantiate a \tbaf $\Omega =\timebipolar$ in order to represent and analyze this example, where:

\begin{itemize}
	\item[] $\ard = \{\argum{A}; \argum{B}; \argum{C}; \argum{D}; \argum{E}; \argum{F}; \argum{G}; \argum{H}; \argum{I}; \argum{J} \}$,
	
	\item[] $\atts = \{(\argum{B},\argum{A}); (\argum{C},\argum{A}); (\argum{H},\argum{F})\}$,
	
	\item[] $\supp = \{(\argum{E},\argum{D}); (\argum{D},\argum{A});(\argum{I},\argum{G}); (\argum{G},\argum{C});(\argum{F},\argum{A}); (\argum{J},\argum{H})\}$, and
	
	\item[] $\av = \{ \tproins{A}{[0-150]}; \tproins{B}{[0-150]}; \tproins{C}{[0-150]}; \tproins{D}{[0-150]};$\linebreak 
	$\tproins{E}{[0-150]}; \tproins{F}{[0-150]}; \tproins{G}{[50-150]}; \tproins{H}{[0-150]};$ \linebreak
	$\tproins{I}{[0-80]}; \tproins{J}{[0-150]}\}$.
\end{itemize}


\begin{figure}[ht]
	\begin{center}\leavevmode
		\xymatrix @R=0pc @C=0pc{
			&{\argu{B}}_{\left[0-150\right]}&&{\argu{I}}_{\left[0-80\right]}&&{\argu{G}}_{\left[50-150\right]}&&{\argu{C}}_{\left[0-150\right]} \\
			&{\blacktriangle}\ar@{..>}[ddddrrrrr]&&{\blacktriangle}\ar@{->}[rr]&&{\blacktriangle}\ar@{->}[rr]&&{\blacktriangle}\ar@{..>}[ddddl] \\
			&&&&&&& \\
			&&&&&&& \\
			&&&&&&& \\
			&{\argu{D}}_{\left[0-150\right]}&{\blacktriangle}\ar@{->}[rrrr]&&&&{\blacktriangle}&{\argu{A}}_{\left[0-150\right]} \\
			&&&&&&& \\
			&&&&&&& \\
			&{\blacktriangle}\ar@{->}[uuur]&&{\blacktriangle}\ar@{->}[rr]&&{\blacktriangle}\ar@{..>}[rr]&&{\blacktriangle}\ar@{->}[uuul] \\
			&{\argu{E}}_{\left[0-150\right]}&&{\argu{J}}_{\left[0-150\right]}&&{\argu{H}}_{\left[0-150\right]}&&{\argu{F}}_{\left[0-150\right]} \\
		}
		\caption{Bipolar argumentation graph with Timed Availability}
		\label{Graph.TimeBipolar2}
	\end{center}
	\vspace*{-15pt}
\end{figure}
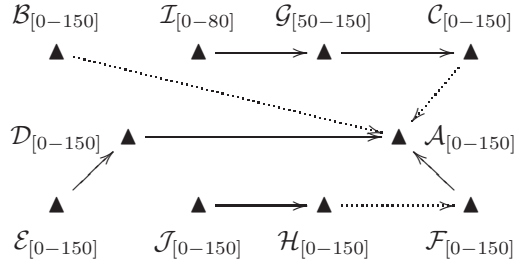

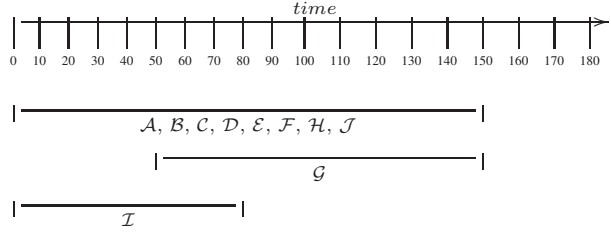
\begin{figure}[ht]
	\begin{center}\leavevmode
		\xymatrix @R=0pc @C=0pc{
			&&&&&&&&&&&&&&&&&&&\\
			\ar@{->}[rrrrrrrrrrrrrrrrrrr]^{time}&&&&&&&&&&&&&&&&&&&\\
			&&&&&&&&&&&&&&&&&&&\\
			\mbox{\tiny 0}\ar@{-}[uuu]&\mbox{\tiny 10}\ar@{-}[uuu]&\mbox{\tiny 20}\ar@{-}[uuu]&\mbox{\tiny 30}\ar@{-}[uuu]&\mbox{\tiny 40}\ar@{-}[uuu]&\mbox{\tiny 50}\ar@{-}[uuu]&\mbox{\tiny 60}\ar@{-}[uuu]&\mbox{\tiny 70}\ar@{-}[uuu]&\mbox{\tiny 80}\ar@{-}[uuu]&\mbox{\tiny 90}\ar@{-}[uuu]&\mbox{\tiny 100}\ar@{-}[uuu]&\mbox{\tiny 110}\ar@{-}[uuu]&\mbox{\tiny 120}\ar@{-}[uuu]&\mbox{\tiny 130}\ar@{-}[uuu]&\mbox{\tiny 140}\ar@{-}[uuu]&\mbox{\tiny 150}\ar@{-}[uuu]&\mbox{\tiny 160}\ar@{-}[uuu]&\mbox{\tiny 170}\ar@{-}[uuu]&\mbox{\tiny 180}\ar@{-}[uuu]\\
			&&&&&&&&&&&&&&&&&&&&\\
			&&&&&&&&&&&&&&&&&&&\\
			\ar@{-}[rrrrrrrrrrrrrrr]_{\argu{A}, \ \argu{B}, \ \argu{C}, \ \argu{D}, \ \argu{E}, \ \argu{F}, \ \argu{H}, \ \argu{J}}&&&&&&&&&&&&&&&&&&&&\\
			\ar@{-}[uu]&&&&&&&&&&&&&&&\ar@{-}[uu]&&&&\\
			&&&&&&&&&&&&&&&&&&&\\
			&&&&&\ar@{-}[rrrrrrrrrr]_{\argu{G}}&&&&&&&&&&&&&&\\
			&&&&&\ar@{-}[uu]&&&&&&&&&&\ar@{-}[uu]&&&&\\
			&&&&&&&&&&&&&&&&&&&\\
			\ar@{-}[rrrrrrrr]_{\argu{I}}&&&&&&&&&&&&&&&&&&&\\
			\ar@{-}[uu]&&&&&&&&\ar@{-}[uu]&&&&&&&&&&&\\
		}
		\caption{Temporal Distribution}
		\label{Graph.DistTime2}
	\end{center}
	\vspace*{-15pt}
\end{figure}

\pagebreak
Let's analyze a couple of collections of t-profiles in order to determine which of them are conflict-free and safe. In one hand, we have the collection {\footnotesize $C_1 =  \{\tproins{C}{[0-80]} ; \linebreak \tproins{G}{(80-150]} ; \tproins{E}{[0-80)} ; \tproins{F}{[0-150]} \}$} which is conflict-free but not safe, since the argument \argum{C} support \argum{A} and \argum{F} attacks \argum{A} in the time interval set {\footnotesize$\{[0-80]\}$}. In another hand, the collection {\footnotesize$C_2 = \{\tproins{C}{[0-80]} ; \tproins{G}{(80-150]} ; \tproins{E}{[0-80)} ; \linebreak \tproins{F}{(80-150]} ; \tproins{A}{[0-80]} \}$} is conflict-free and safe, since the argument \argum{C} that support \argum{A} is available when \argum{F} (that attacks \argum{A}) is not (\argum{C} is available in the time interval {\footnotesize$[0-80]$}, while \argum{F} is available in the interval {\footnotesize$(80-150]$}).  

Let's determine the collection of t-profile that represent a t-stable extension. In this case, the collection {\footnotesize$C_3 = \{\tproins{C}{[0-80]} ;  \tproins{G}{(80-150]} ; \tproins{E}{[0-150]} ; \linebreak \tproins{A}{[0-80]} ; \tproins{I}{[0-80]} ; \tproins{J}{[0-150]} ; \tproins{B}{[0-150]} \}$} is a safe t-stable extension, since its conflict-free and attacks all the t-profiles not considered in $C_3$. In this case these t-profiles are: \\

$ \begin{array}{ll}
\tproins{C}{(80-150]} & \mbox{support defeated by } \tproins{G}{[80-150]} \\
\tproins{G}{[50-80]} & \mbox{support defeated by } \tproins{I}{[0-80]} \\
\tproins{D}{[0-50]} & \mbox{support defeated by } \tproins{E}{[0-150]} \\
\tproins{H}{[0-150]} & \mbox{support defeated by } \tproins{J}{[0-150]} \\
\tproins{F}{[0-150]} & \mbox{secondary defeated by } \tproins{J}{[0-150]} \mbox{ and } \tproins{H}{[0-150]} \\
\tproins{A}{(80-150]} & \mbox{secondary defeated by } \tproins{G}{[80-150]} \mbox{ and } \tproins{C}{(80-150]} \\  	    	  
\end{array}$\\
\\
\

Finally, using the results obtained in \emph{Proposition} 3 that relates  by t-inclusion the td-preferred, ts-preferred and tc-preferred extensions we can conclude that the collection of t-profiles $C_3$, which is a safe t-stable, is also td-preferred, ts-preferred and tc-preferred. 
We only need to show that it is td-preferred. This means that $C_3$ should be maximal and td-acceptable (\ie\ conflict-free and must defends all its elements).  We have already shown that $C_3$ is conflict-free and by attacking all the t-profiles that do not belong to $C_3$ we can assure its maximality in acceptability. 

In this particular case, $C_3$ is a safe t-stable, td-preferred, ts-preferred and tc-preferred extension. This situation occurs because the bipolar argumentation $\Omega$ do not include cycles. The graph representing our example is acyclic \textit{in every moment of time}. As we see in the abstract example, the difference  is produced by cycles of support and attacks, as it was proved elsewhere for classical, non bipolar argumentation frameworks 

\section{Related Work}\label{sec.RelatedWork}

As discussed in the introduction, reasoning about time is an important concern in commonsense reasoning. Thus, its consideration becomes relevant when modeling argumentation capabilities of intelligent agents~\cite{rahwan2009argumentation}. There have been recent advances in modeling time in argumentation frameworks. Mann and Hunter in~\cite{MannHunterComma}, propose a calculus for representing temporal knowledge, which is defined in terms of propositional logic. The use of this calculus is then considered with respect to argumentation, where an argument is defined in the standard way: an argument is a pair constituted by a minimally consistent subset of a database entailing its conclusion. Briefly speaking, the authors discuss a way of encoding temporal information into propositional logic, and examined its impact in a coherence argumentation system. The central idea is that they draw heavily on temporal knowledge due to their day-to-day nature - what is true today may well not be true tomorrow - as well as their inclusion of information concerning periods of time. In order to represent time variable, the authors propose a calculus built upon the ideas of Allen’s interval logic~\cite{Allen83} using abstract intervals, and in keeping with the desire for a practical system, they restrict the system to using specific timepoints. This work is related to the works proposed by Hunter in~\cite{hunter2001ramification} and Augusto and Simari in~\cite{AugustoSimari01}. Hunter’s system is based on maximally consistent subsets of the knowledge base, which are now not normally regarded as representative of arguments, while Augusto and Simari’s contribution is based upon a many sorted logic with defeasible formula, and hence also falls into a different category of argumentation, and the use of many sorted logic raises similar concerns to that of using first order logic. 

Barringer \emph{et.al.} present two important approaches that share elements of our research. In the first one~\cite{barringer2010modal}, the authors present a temporal argumentation approach, where they extend the traditional Dung’s networks using temporal and modal language formulas to represent the structure of arguments. To do that, they use the concept of usability of arguments defined as a function that determines if an argument is usable or not in a given context, changing this status over time based on the change in a dynamics context. In addition, they improved the representational capability of the formalism by using the ability of modal logic to represent accessibility between different argumentative networks; in this way, the modal operator is treated as a fibring operator to obtain a result for another argumentation network context, and then apply it to the local argumentation network context. In the second~\cite{barringer2012temporal}, they study the relationships of support and attack between arguments through a numerical argumentation network, where both the strength of the arguments and the strength that carry the attack and support between them is considered. This work pays close attention to the relations of support and attack between arguments, and to the treatment of cycles in an argumentative network. Furthermore, they offer different motivations for modeling domains in which the strengths can be time-dependent, presenting a brief explanation of how to deal with this issue in a numerical argumentation network.

Finally, Godo and Pardo~\emph{et.al.} in~\cite{pardo2011t} and Bud\'an~\emph{et.al.} in~\cite{budan2012approach}, explored the possibility of expressing the uncertainty or reliability of temporal rules and events, and how this features may change over time. In the first one~\cite{pardo2011t}, the authors propose an argumentation-based defeasible logic, called \emph{t-DeLP}, that focuses on forward temporal reasoning for causal inference. They extend the language of the \emph{DeLP} logical framework by associating temporal parameters to literals. As usual, a dialectical procedure determines which arguments are undefeated, and hence which literals are warranted, or defeasibly follow from the program. \emph{t-DeLP}, though, slightly differs from \emph{DeLP} in order to accommodate temporal aspects, like the persistence of facts. The output of a t-DeLP program is a set of warranted literals, which is first shown to be non-contradictory and be closed under sub-arguments. This basic framework is then modified to deal with programs whose strict rules encode mutex constraints. The resulting framework is shown to satisfy stronger logical properties like indirect consistency and closure. In the second~\cite{budan2012approach},  the authors present a different extension of \emph{DeLP} introducing the possibility of formalizing arguments and the corresponding defeat relations among them by combining both temporal criteria and belief strength criteria. This extension is based on the \emph{Extended Temporal Argumentation Framework} (\emph{E-TAF})~\cite{cobo2010admissibility,budan2015modeling} which has the capability of modeling different time-dependent properties associated with arguments. Briefly speaking, this extension of \emph{DeLP} incorporate the representation of temporal availability and strength factors of arguments varying overtime, associating these characteristics with the DeLP language elements. The strength factors are used to model different more concrete measures such
as reliability, priorities, etc.; the information is propagated to the level of arguments, and then the \emph{E-TAF} definitions are applied establishing their temporal acceptability 

Analyzing these research lines, is quite dificult to establish a proper comparison among the works mentioned above and \tbaf. This complication arises from the different levels of abstraction used, some of them are not abstract formalisms while other uses other temporal represententation (based on events or modalities). There is a clear relation with non-temporal approaches \cite{CohenGGS14}, this aspects are explored through out the paper.

\section{Conclusions and Future Work}\label{sec.conclu}

In this work we expanded temporal argumentation frameworks (\emph{TAF}) to include an argument support relation, as in classical bipolar argumentation frameworks. In this formalization, arguments are only valid for consideration (\textit{available} or \textit{relevant}) in a given period of time, which is defined for every individual argument. Hence, support and defeat relation are sporadic and proper argument semantics are defined. We bring admissibility-based extensions for bipolar scenarios to the context of timed argumentation, providing new formalizations of argument semantics with time involved.

Future work has several directions. We view temporal information as an additional dimension that can be applied to several argumentation models. We are interested in the formalization of other timed argument relations, specially the ones defined in the backing-undercutting argumentation framework of \cite{CohenGS11}. Also, we will investigate how the approach could be developed by considering a timed version of Caminada’s labelling, where an argument has a particular label for a specified period of time. Besides interval-based semantics defined in this present work, we are also interested in new integrations of timed notions in argumentation, such as temporal modal logic~\cite{gabbay2003many,barringer2012temporal}. We are developing of a framework combining the representation capabilities of \baf with an algebra of argumentation labels \cite{budanWL4AI13} to represent timed features of arguments in dynamic domains.

\bibliographystyle{elsarticle-num}

\bibliography{bib}

\appendix
\section{Proofs}
\setcounter{Proposition}{0}
\setcounter{Lemma}{0}
\setcounter{Theorem}{0}

\begin{Proposition}
	Let $S$ be a collection of t-profiles:
	\begin{itemize} 
		\item[--] If $S$ is safe, then any collection $S' \subseteq_t S$ is conflict-free. 
		\item[--] If $S$ is conflict-free and closed for \supp then $S$ is safe.
	\end{itemize}
\end{Proposition}

\noindent \underline{Proof:}  We will separate the proof in two parts according to the statements giving in the proposition.

\begin{itemize}
	\item[--] \emph{If $S$ is safe, then any collection $S' \subseteq_t S$ is conflict-free}: If $C$ is safe then, for definition, it can't be the case that a t-profile $\tpro{A}$, that belong to the collection $C$, be supported for any t-profile $\tpro{C}$ that belong to $C$ and defeat (by a supported or secondary defeat) a t-profile $\tpro{B} \in C$, both at the same time. More formally  $\nexists \tpro{A}, \tpro{B} \in C$ and  $\nexists \tpro{B}, \tpro{C} \in C$ such that \tpro{A} supported defeat \tpro{B} with a $\tsuppdefeat{A}{B} \neq \emptyset$ or \tpro{A} secondary defeat \tpro{B} with a $\tsecdefeat{A}{B} \neq \emptyset$, and either there is a sequence of support from \tpro{C} to \tpro{A}, or $\tpro{A} \in C$. 
	Suppose that there is a collection $C' \subseteq_t C$ that is not conflict-free. Then, $\exists \tpro{A}, \tpro{B} \in C'$ such that \tpro{A} supported defeat \tpro{B} with a $\tsuppdefeat{A}{B} \neq \emptyset$ or \tpro{A} secondary defeat \tpro{B} with a $\tsecdefeat{A}{B} \neq \emptyset$. This lead us to a contradiction, since $C' \subseteq_t C$ defining that for any t-profile $\tproprima{X} \in C'$ there exists a t-profile $\tpro{X} \in C$ such that $\tiempoprima{X}  \subseteq \tiempo{X}$ where $\tsuppdefeat{A}{B} = \emptyset$ and $\tsecdefeat{A}{B} = \emptyset$ for pairs of t-profile in $C$. 
	\item[--] \emph{If $S$ is conflict-free and closed for \supp then $S$ is safe}: Since for hypothesis the collection of t-profiles $C$ is conflict-free and closed for \supp, then there is not exist a t-profile $(\argum{X},\tdefencese{X}) \in C$, a t-profile $(\argum{Y},\tdefencese{Y}) \in C$ and a t-profile $(\argum{Z},\tiempo{Z} \setminus \tdefencese{Z})$ such that $(\argum{Y},\tdefencese{Y})$ has a support sequence from $(\argum{Z},\tiempo{Z} \setminus \tdefencese{Z})$ in the time intervals $\tdefencese{Y} \cap (\tiempo{Z} \setminus \tdefencese{Z})$, and there should not exist attacks (secondary or supported defeat) from $(\argum{Z},\tiempo{Z} \setminus \tdefencese{Z})$ to $(\argum{X},\tdefencese{X})$ in the time interval $\tdefencese{X} \cap \tdefencese{Y} \cap (\tiempo{Z} \setminus \tdefencese{Z})$. So, $C$ maintains an internal and external coherence satisfying the safe property.
\end{itemize}	

\begin{Proposition}
	Let $\Omega= \langle \ard,$ $ \atts, \supp,\av \rangle$ be a \tbaf, then:
	\begin{itemize} 
		\item[--] A td-admissible extension is t-included in a ts-admissible extension.
		
		\item[--] A ts-admissible extension is t-included in a tc-admissible extension.
	\end{itemize}
\end{Proposition}

\noindent \underline{Proof:}  We will separate the proof in two parts according to the statements giving in the proposition.

\begin{itemize}
	\item[--] \emph{A td-admissible extension is t-included in a ts-admissible extension}: Let suppose there is a collection of t-profiles $C$ such that is ts-admissible but not td-admissible. By definition if a collection of t-profile $C$ is a ts-admissible extension, then $C$ is safe and defend all its elements. If $C$ is no td-admissible then is not conflict-free or it fails in defending all its elements. This lead us to a contradiction that arises from the supposition. $C$ can't defend and fail defending all its elements at the same time, and cannot be safe and not conflict-free (see Proposition 1). So all collection of t-profiles that is td-admissible is also ts-admissible.
	
	\item[--] \emph{A ts-admissible extension is t-included in a tc-admissible extension:} Let supose there is a collection of t-profiles $C$ such that is tc-admissible by no ts-admissible. By definition if a collection of t-profile $C$ is a tc-admissible extension, then $C$ is conflict-free, closed under \supp, and defend all its elements. If $C$ is no ts-admissible then is not safe or it fails in defending all its elements. This lead us to a contradiction that arises from the supposition. $C$ can't defend and fail defending all its elements at the same time, and cannot be safe and conflict-free and closed under \supp (see Proposition 1). So all collection of t-profiles that is tc-admissible is also ts-admissible.
\end{itemize}

\begin{Proposition}
	Let $\Omega= \langle \ard,$ $ \atts, \supp,\av \rangle$ be a \tbaf, then:
	
	\begin{itemize}
		\item[-] A td-preferred extension is t-included in a ts-preferred extension.
		
		\item[-] A ts-preferred extension is t-included in a tc-preferred extension.
		
		\item[-] A ts-preferred extension closed under \supp is also a tc-preferred.
		
		\item[-] A td-preferred extension is t-included in a t-stable extension.
		
		\item[-] A ts-preferred extension is t-included in a safe t-stable extension.
		
		\item[-] A tc-preferred extension is t-included in a safe t-stable extension.
	\end{itemize}
\end{Proposition}

\noindent\underline{Proof:} We will separate the proof in the six parts of the proposition corresponding to each of the six relations.\\

\begin{itemize}
	\item[i)] \emph{A td-preferred extension is t-included in a ts-preferred extension}: Lets assume that $C$ is a ts-preferred extension, $C'$ is a td-preferred extension, and  $C' \varsubsetneq_t C$. However, $C$ is maximal w.r.t. set t-inclusion among the ts-admissible, while $C'$ is maximal w.r.t. set t-inclusion among the td-admissible. In addition, by \emph{Proposition} 2, we know that td-admissible extension is t-included in a ts-admissible extension. So, this contradicts the assumption that $C$ is a ts-preferred extention or that it exists $C' \varsubsetneq_t C$ with $C'$ being a td-preferred extension. So a td-preferred extension is t-included in a ts-preferred extension.
	\item[ii)] \emph{A ts-preferred extension is t-included in a tc-preferred extension}: Lets assume that $C$ is a tc-preferred extension, $C'$ is a ts-preferred extension, and  $C' \varsubsetneq_t C$. However, $C$ is maximal w.r.t. set t-inclusion among the tc-admissible, while $C'$ is maximal w.r.t. set t-inclusion among the ts-admissible. In addition, by \emph{Proposition} 2, we know that ts-admissible extension is t-included in a tc-admissible extension. So, this contradicts the assumption that $C$ is a tc-preferred extention or that it exists $C' \varsubsetneq_t C$ with $C'$ being a ts-preferred extension. So a ts-preferred extension is t-included in a tc-preferred extension.
	\item[iii)] \emph{A ts-preferred extension cloused under \supp is also a tc-preferred}: Lets assume that $C$ is a ts-preferred extension closed under \supp. This means that, $C$ is safe, is closed under \supp, and defends all its elements. In addition, we know that if $C$ is safe, then $C$ is conflict free by \emph{Proposition} 1. In this sense, $C$ es conflict free, closed under support and defends all its elements corresponding with the tc-admissible definition. In addition, $C$ is the maximal set w.r.t. set inclusion. Consequently, $C$ is a tc-preferred extension.  
	\item[iv)] \emph{A td-preferred extension is t-included in a t-stable extension}: Lets assume that $C$ is a t-stable extension, $C'$ is a td-preferred extension, and  $C' \varsubsetneq_t C$. Then, there should exist at least a t-profile $(\argum{X},\mathcal{T}_{(\argum{X} \mid C')}) \in C'$ and a t-profile $(\argum{X},\mathcal{T}_{(\argum{X} \mid C)}) \in C$ such that $\mathcal{T}_{(\argum{X} \mid C')} \varsubsetneq  \mathcal{T}_{(\argum{X} \mid C)}$. In particular, there exist a time point $\alpha$ verifying that $\alpha \in \mathcal{T}_{(\argum{X} \mid C')}$ and $\alpha \notin \mathcal{T}_{(\argum{X} \mid C)}$. In this sense, there exist a time point $\alpha$ where an argument $\argum{X}$ is defended in $C'$ and not in $C$. Contradiction, $C$ is a stable extension defeating all the t-profiles that not belong to it. Then, if the argument $\argum{X}$ is defended by $C'$ in $\alpha$, it is also defended by $C$ in such time point.
	\item[v)] \emph{A ts-preferred extension is t-included in a safe t-stable extension}: For item iii) we establish that all td-preferred extension is t-included in a ts-preferred extension, and for i) all td-preferred extension is t-included in a t-stable extension. Then, the only way the t-inclusion can fail is by the existence of a ts-prefered extension, that is not a td-preferred extension, that is not a safe t-stable extension. By definition of ts-preferred extension and td-preferred extension the condition the only condition that marks difference is that the first one grants safety. Lets suppose there is a set $C$ that is a ts-preferred extension that is not a safe t-stable extension, this assumption lead us inmediatly to contradicition since the safety condition is granted in both sets. 
	%
	\item[vi)] \emph{A tc-preferred extension is t-included in a safe t-stable extension}: For item ii) we establish that all ts-preferred extension is t-included in a tc-preferred extension, and for iv) all ts-preferred extension is t-included in a safe t-stable extension. The only difference with item v) is that a tc-preferred extension grants closed under \supp. However, for \emph{Proposition} 1, if a conllection of t-profile $C'$ is conflict-free and closed for \supp, then $C'$ is safe. In addition, by hypothesis the t-stable extension satisfy safety condition. So, A tc-preferred extension is t-included in a safe t-stable extension.
	
\end{itemize}

\begin{Theorem}
	Let $\Omega= \timebipolar$ be a \tbaf and let $\alpha$ representing a point in time. Let $\Theta'_{\alpha} = \langle \ard'_{\alpha}, \atts^{\alpha}, \supp^{\alpha} \rangle$ be a
	bipolar abstract framework obtained from $\Omega$ in the following way: $\ard'_{\alpha} = \{\argum{A} \in \ard \mid \alpha \in \tiempo{A}\}$,  $\atts^{\alpha} = \{(\argum{A}, \argum{B})\in \alpha \in \timedefeat{A}{B}\}$ and $\supp^{\alpha} = \{(\argum{A}, \argum{B})\in \alpha \in \timesupport{A}{B}\}$.
	Let $C$ a collection of t-profiles in $\Omega$, and $C'_{\alpha} = \{\argum{A} \mid  \tdefencese{A} \in C$ and $\alpha \in \tdefencese{A}\}$. It holds that, if $C$ is an td-preferred extension (resp. ts-preferred, tc-preferred, and t-stable) w.r.t. $\Omega$, then $C'_{\alpha}$ is a  d-preferred extension (resp. ts-preferred, tc-preferred, and t-stable) w.r.t. $\Theta'_{\alpha}$.
\end{Theorem} 

\noindent\underline{Proof:}  We will separate the proof in the four parts of the theorem corresponding to each of the four semantics. First, we will prove the general property of conflict-freeness that any extension corresponding to any semantics should satisfy (the safe property require that a collection of t-profile has the conflict-free property):\\

\noindent If $C$ is an extension w.r.t. $\Omega$, then $C'_{\alpha}$ is should be conflict-free w.r.t. $\Theta'_{\alpha}$.\\

\noindent Let us assume that $C'_{\alpha}$ is not a conflict-free set of arguments. In that case, there should exist two arguments $X,Y \in C'_{\alpha}$ such that $X$ support defeat $Y$ or $X$ secondary defeat $Y$ through a sequence of arguments $\argum{X} \ \R_1 \ ... \ \R_n \ \argum{Y}$. From the definition of $C'_{\alpha}$, we know that there are $(X,\tdefencese{X}),(Y,\tdefencese{Y}) \in C$, such that $\alpha \in \tdefencese{X} \cap \tdefencese{Y}$, and $\alpha \in \tsuppdefeat{X}{Y} \cup \tsecdefeat{X}{Y}$. Consequently, $C$ is not an at-conflict-free set contradicting our initial assumption that $C$ is an extension, and this contradiction comes from assuming that $C'_{\alpha}$ is not a conflict-free set.\\ 

\noindent We will now proceed under the assumption that $C'_{\alpha}$ is a conflict-free set for the four semantics mentioned.\\

\noindent a) If $C$ is a td-preferred extension w.r.t. $\Omega$, then $C'_{\alpha}$ is a d-preferred extension w.r.t. $\Theta'_{\alpha}$.\\

\noindent Let $C$ be a td-preferred extension for $\Omega$, and let $C'_{\alpha}$ be a set of arguments such that $C'_{\alpha} = \{ \argum{X} \mid (\argum{X},\tdefencese{X}) ) \in C \ \text{and} \ \alpha \in \tdefencese{X}\}$, and let us assume that $C'_{\alpha}$ is not a d-preferred extension of $\Theta'_{\alpha}$. For this to be the case, knowing $C'_{\alpha}$ is conflict-free, at least one of the two conditions required for d-preferred semantics should fail, namely:

\begin{itemize}
	\item[i)] $C'_{\alpha}$	should be a d-admissible set. Let us assume that $C'_{\alpha}$ does not satisfy	that condition. In this case, there should exist two arguments $\argum{X},\argum{Y} \in \ard'_{\alpha}$, such that $\argum{Y} \notin C'_{\alpha}$, $\argum{X} \in C'_{\alpha}$,  $\argum{Y}$ support defeat $\argum{X}$ or $\argum{Y}$ secondary defeat $\argum{X}$, and there should not exist an argument $\argum{Z} \in \ard'_{\alpha}$ verifying that $\argum{Z} \in C'_{\alpha}$ and $\argum{Z}$ support defeat $\argum{Y}$ or $\argum{Z}$ secondary defeat $\argum{Y}$. From the definition of $C'_{\alpha}$, we know that there is a t-profile $(\argum{X},\tdefencese{X}) \in C$ where $\alpha \in \tdefencese{X}$, a t-profile $(\argum{Y},\tiempo{Y})$ such that $\tsuppdefeat{Y}{X} \cup \tsecdefeat{Y}{X} \neq \emptyset$ with $\alpha \in \tsuppdefeat{Y}{X} \cup \tsecdefeat{Y}{X}$, and does not exist a t-profile $(\argum{Z},\tdefencese{Z}) \in C$ such that $\tsuppdefeat{Z}{Y} \cup \tsecdefeat{Z}{Y} \neq \emptyset$ with $\alpha \in \tsuppdefeat{Z}{Y} \cup \tsecdefeat{Z}{Y}$ verifying that $(\tsuppdefeat{Y}{X} \cup \tsecdefeat{Y}{X}) \cap (\tsuppdefeat{Z}{Y} \cup \tsecdefeat{Z}{Y}) \neq \emptyset$. But $C$ is a td-preferred extension, and therefore it should satisfies dt-admissibility, in particular that $C$ defends all its elements in the time intervals in which each t-profiles belong to it (contradiction).
	\item[ii)] $C'_{\alpha}$ should be a maximal set w.r.t. inclusion. Let us assume that is not, then there exists a set $C''_{\alpha}$ such that $C'_{\alpha} \subsetneq C''_{\alpha}$ and it satisfies conflict-freeness and admissibility. Let $C_m = C \cup \{(\argum{X},{\alpha}) \mid \argum{X} \in C''_{\alpha}$ and $\argum{X} \notin C'_{\alpha}\}$. Note that $C \subsetneq_t C_m$ (by construction). Also, $C_m$ is td-admissible. Contradiction, since $C$ is an td-preferred extension and therefore it is the maximal collection of t-profiles w.r.t. t-inclusion which is td-admissible. $\Box$
\end{itemize}

\noindent b) If $C$ is a ts-preferred extension w.r.t. $\Omega$, then $C'_{\alpha}$ is a s-preferred extension w.r.t. $\Theta'_{\alpha}$.\\

\noindent Let $C$ be a ts-preferred extension for $\Omega$, and let $C'_{\alpha}$ be a set of arguments such that $C'_{\alpha} = \{ \argum{X} \mid (\argum{X},\tdefencese{X}) ) \in C \ \text{and} \ \alpha \in \tdefencese{X}\}$, and let us assume that $C'_{\alpha}$ is not a s-preferred extension of $\Theta'_{\alpha}$. For this to be the case, knowing $C'_{\alpha}$ is conflict-free, at least one of the two conditions required for s-preferred semantics should fail, namely:

\begin{itemize}
	\item[i)] $C'_{\alpha}$	should be a s-admissible set. Let us assume that $C'_{\alpha}$ does not satisfy	that condition. In this case, two situations can arise: (a) there should exist two arguments $\argum{X},\argum{Y} \in \ard'_{\alpha}$, such that $\argum{Y} \notin C'_{\alpha}$, $\argum{X} \in C'_{\alpha}$,  $\argum{Y}$ support defeat $\argum{X}$ or $\argum{Y}$ secondary defeat $\argum{X}$, and there should not exist an argument $\argum{Z} \in \ard'_{\alpha}$ verifying that $\argum{Z} \in C'_{\alpha}$ and $\argum{Z}$ support defeat $\argum{Y}$ or $\argum{Z}$ secondary defeat $\argum{Y}$. From the definition of $C'_{\alpha}$, we know that there is a t-profile $(\argum{X},\tdefencese{X}) \in C$ where $\alpha \in \tdefencese{X}$, a t-profile $(\argum{Y},\tiempo{Y})$ such that $\tsuppdefeat{Y}{X} \cup \tsecdefeat{Y}{X} \neq \emptyset$ and $\alpha \in \tsuppdefeat{Y}{X} \cup \tsecdefeat{Y}{X}$, and does not exist a t-profile $(\argum{Z},\tdefencese{Z}) \in C$ such that $\tsuppdefeat{Z}{Y} \cup \tsecdefeat{Z}{Y} \neq \emptyset$ with $\alpha \in \tsuppdefeat{Z}{Y} \cup \tsecdefeat{Z}{Y}$ verifying that $(\tsuppdefeat{Y}{X} \cup \tsecdefeat{Y}{X}) \cap (\tsuppdefeat{Z}{Y} \cup \tsecdefeat{Z}{Y}) \neq \emptyset$. But $C$ is a td-preferred extension, and therefore it should satisfies st-admissibility, in particular that $C$ defends all its elements in the time intervals in which each t-profiles belong to it (contradiction); 
	or (b) there should exist three arguments $\argum{X},\argum{Y},\argum{Z} \in \ard'_{\alpha}$, such that $\argum{Z} \notin C'_{\alpha}$, $\argum{X} \in C'_{\alpha}$, $\argum{Y} \in C'_{\alpha}$, $\argum{X}$ support defeat $\argum{Z}$ or $\argum{X}$ secondary defeat $\argum{Z}$, and there exist a sequence of support from $\argum{Y}$ to $\argum{Z}$. From the definition of $C'_{\alpha}$, we know that there is a t-profile $(\argum{X},\tdefencese{X}) \in C$ where $\alpha \in \tdefencese{X}$, a t-profile $(\argum{Y},\tdefencese{Y}) \in C$ where $\alpha \in \tdefencese{Y}$, and a t-profile $(\argum{Z},\tiempo{Z})$ such that $\tsuppdefeat{X}{Z} \cup \tsecdefeat{X}{Z} \neq \emptyset$ with $\alpha \in \tsuppdefeat{X}{Z} \cup \tsecdefeat{X}{Z}$ and  $\tdefencese{Y} \cap (\tsuppdefeat{X}{Z} \cup \tsecdefeat{X}{Z}) \neq \emptyset$ with $\alpha \in \tdefencese{Y} \cap (\tsuppdefeat{X}{Z} \cup \tsecdefeat{X}{Z})$. But $C$ is a td-preferred extension, and therefore it should satisfies st-admissibility, in particular the safe condition (contradiction);   
	\item[ii)] $C'_{\alpha}$ should be a maximal set w.r.t. inclusion. Let us assume that is not, then there exists a set $C''_{\alpha}$ such that $C'_{\alpha} \subsetneq C''_{\alpha}$ and it satisfies conflict-freeness and admissibility. Let $C_m = C \cup \{(\argum{X},{\alpha}) \mid \argum{X} \in C''_{\alpha}$ and $\argum{X} \notin C'_{\alpha}\}$. Note that $C \subsetneq_t C_m$ (by construction). Also, $C_m$ is td-admissible. Contradiction, since $C$ is an td-preferred extension and therefore it is the maximal collection of t-profiles w.r.t. t-inclusion which is td-admissible. $\Box$
\end{itemize}

\noindent c) If $C$ is a tc-preferred extension w.r.t. $\Omega$, then $C'_{\alpha}$ is a c-preferred extension w.r.t. $\Theta'_{\alpha}$.\\

\noindent Let $C$ be a tc-preferred extension for $\Omega$, and let $C'_{\alpha}$ be a set of arguments such that $C'_{\alpha} = \{ \argum{X} \mid (\argum{X},\tdefencese{X}) ) \in C \ \text{and} \ \alpha \in \tdefencese{X}\}$, and let us assume that $C'_{\alpha}$ is not a c-preferred extension of $\Theta'_{\alpha}$. For this to be the case, knowing $C'_{\alpha}$ is conflict-free, at least one of the three conditions required for c-preferred semantics should fail, namely:

\begin{itemize}
	\item[i)] $C'_{\alpha}$	should be a c-admissible set. Let us assume that $C'_{\alpha}$ does not satisfy	that condition. In this case, two situations can arise: (a) there should exist two arguments $\argum{X},\argum{Y} \in \ard'_{\alpha}$, such that $\argum{Y} \notin C'_{\alpha}$, $\argum{X} \in C'_{\alpha}$,  $\argum{Y}$ support defeat $\argum{X}$ or $\argum{Y}$ secondary defeat $\argum{X}$, and there should not exist an argument $\argum{Z} \in \ard'_{\alpha}$ verifying that $\argum{Z} \in C'_{\alpha}$ and $\argum{Z}$ support defeat $\argum{Y}$ or $\argum{Z}$ secondary defeat $\argum{Y}$. From the definition of $C'_{\alpha}$, we know that there is a t-profile $(\argum{X},\tdefencese{X}) \in C$ where $\alpha \in \tdefencese{X}$, a t-profile $(\argum{Y},\tiempo{Y})$ such that $\tsuppdefeat{Y}{X} \cup \tsecdefeat{Y}{X} \neq \emptyset$ with $\alpha \in \tsuppdefeat{Y}{X} \cup \tsecdefeat{Y}{X}$, and does not exist a t-profile $(\argum{Z},\tdefencese{Z}) \in C$ such that $\tsuppdefeat{Z}{Y} \cup \tsecdefeat{Z}{Y} \neq \emptyset$ with $\alpha \in \tsuppdefeat{Z}{Y} \cup \tsecdefeat{Z}{Y}$ verifying that $(\tsuppdefeat{Y}{X} \cup \tsecdefeat{Y}{X}) \cap (\tsuppdefeat{Z}{Y} \cup \tsecdefeat{Z}{Y}) \neq \emptyset$. But $C$ is a tc-preferred extension, and therefore it should satisfies tc-admissibility, in particular that $C$ defends all its elements in the time intervals in which each t-profiles belong to it (contradiction); or (b) there should exist two arguments $\argum{X},\argum{Y} \in \ard'_{\alpha}$, such that there exist a support sequence from $\argum{X}$ to $\argum{Y}$, $\argum{Y} \notin C'_{\alpha}$, $\argum{X} \in C'_{\alpha}$, and not exist an argument $\argum{Z} \in \ard'_{\alpha}$ verifying that $\argum{Z}$ support defeat $\argum{Y}$ or $\argum{Z}$ secondary defeat $\argum{Y}$. From the definition of $C'_{\alpha}$, we know that there is a t-profile $(\argum{X},\tdefencese{X}) \in C$ where $\alpha \in \tdefencese{X}$ and a t-profile $(\argum{Y},\tiempo{Y})$ such that $\tdefencese{X}\cap \tiempo{Y} \neq \emptyset$ with $\alpha \in \tdefencese{X}\cap \tiempo{Y}$. But $C$ is a tc-preferred extension, and therefore it should satisfies tc-admissibility, in particular that $C$ is closed under \supp (contradiction).
	\item[ii)] $C'_{\alpha}$ should be a maximal set w.r.t. inclusion. Let us assume that is not, then there exists a set $C''_{\alpha}$ such that $C'_{\alpha} \subsetneq C''_{\alpha}$ and it satisfies conflict-freeness and admissibility. Let $C_m = C \cup \{(\argum{X},{\alpha}) \mid \argum{X} \in C''_{\alpha}$ and $\argum{X} \notin C'_{\alpha}\}$. Note that $C \subsetneq_t C_m$ (by construction). Also, $C_m$ is td-admissible. Contradiction, since $C$ is an td-preferred extension and therefore it is the maximal collection of t-profiles w.r.t. t-inclusion which is td-admissible. $\Box$
\end{itemize}

\noindent d) If $C$ is a ts-stable extension w.r.t. $\Omega$, then $C'_{\alpha}$ is a stable extension w.r.t. $\Theta'_{\alpha}$.\\

\noindent Let $C$ be a t-stable extension for $\Omega$, and let $C'_{\alpha}$ be a set of arguments such that $C'_{\alpha} = \{ \argum{X} \mid (\argum{X},\tiempo{X}) ) \in C \ \text{and} \ \alpha \in \tiempo{X}\}$, and let us assume that $C'_{\alpha}$ is not a stable extension of$\Theta'_{\alpha}$. For this to be the case, knowing $C'_{\alpha}$ is conflict-free, the following condition required for stable semantics should fail: $C'_{\alpha}$ should attack (secondary or support defeat) all arguments that do not belong to it. Let us assume that the condition fails; then, there exist at least an argument $\argum{X} \in \ard'_{\alpha} \setminus C'_{\alpha}$ that is not attacked (secondary or support defeat) by any argument in  $C'_{\alpha}$. Consequently, there exists a t-profile $(\argum{X},\tiempo{X}) \notin C$, where $C = \{(\argum{Y}_i, \mathcal{T}_{(\argum{Y}_i,C)}\mid 1 \leq i \leq n\}$ such that $\alpha \in \tiempo{X}$ and therefore and therefore $\tiempo{X} \setminus \bigcup^n_{i=1}  \mathcal{T}_{(\argum{Y}_i,C)} \neq \emptyset$ since it contains at least the time point $\alpha$. But this is not possible since $C$ is an t-stable extension, thus $C$ attacks all the t-profiles that do not belong to that t-stable extension, in particular this is true for $(\argum{X}, \tiempo{X})$.	This is a contradiction that arises from our assumption that $C'_{\alpha}$ does not attack all arguments that are outside of it.

\begin{Proposition}
	Let $\Omega= \langle \ard,$ $ \atts, \supp,\av \rangle$ be a \tbaf, and \tpro{A} and \tpro{B} be two t-profiles, where \argum{B} defeats \argum{A} through a support or secondary attacks, then it holds that:
	\begin{itemize}
		\item[--] $t\mbox{-}ES(\argum{A}) \cap t\mbox{-}ES(\argum{B}) = \emptyset$
		\item[--] $t\mbox{-}PR_d(\argum{A}) \cap t\mbox{-}PR_d(\argum{B}) = \emptyset$;
		\item[--] $t\mbox{-}PR_s(\argum{A}) \cap t\mbox{-}PR_s(\argum{B}) = \emptyset$;
		\item[--] $t\mbox{-}PR_c(\argum{A}) \cap t\mbox{-}PR_c(\argum{B}) = \emptyset$; and
	\end{itemize}
\end{Proposition}

\noindent\underline{Proof:}  Let $C$ be the t-stable extension for $\Omega$. Let \tpro{A}, \tpro{B} be two t-profiles in $C$. Since $C$ is the t-stable extension for $\Omega$, then $C$ is conflict-free and for all $\tpro{A} \notin C$, verifies that $\tiempo{A} \ \setminus \ (\bigcup \tsecdefeat{B}{A} \ \cup \ \bigcup \tsuppdefeat{B}{A}) = \emptyset$ for all $\tpro{B} \in C$. For definition of conflict-free there is no t-profiles $\tpro{A},\tpro{B} \in C$ such that $(\tpro{A},\tpro{B}) \in \atts$ and $\tsuppdefeat{A}{B} \neq \emptyset$ or $\tsecdefeat{A}{B} \neq \emptyset$. Therefore, $t\mbox{-}ES(\argum{A}) \cap t\mbox{-}ES(\argum{B}) = \emptyset$.

The proof of the result is based on the property that states that the collection $C$ is conflict free. This is a implicit requirement for the t-stable extension, the tc-preferred extension, and the td-preferred through the notion of td-admissible ($C$ is td-admissible if $C$ is conflict-free and defends all its elements), meaning there cannot be a conflict between two elements of $C$. In another hand, the ts-preferred extension established that the collection $C$ must satisfy the internal and external coherence, satisfying the conflict-free condition ( $C$ is \emph{Safe} iff $\nexists \tpro{A}, \tpro{B} \in S$ and  $\nexists \tpro{C}$ where $\tpro{C}$ is a valid $\Omega$'s t-profile such that $\tsuppdefeat{A}{C} \neq \emptyset$ or $\tsecdefeat{A}{C} \neq \emptyset$, and either there is a sequence of support from \tpro{B} to \tpro{C}, or $\tpro{C} \in C$). Consequently, the proof for the other three extensions are analogous.$\Box$


\end{document}

%% file: Comandos.tex
\newtheorem{Definition}{\textbf{Definition}}

\newtheorem{Theorem}{\textbf{Theorem}}

\newtheorem{Proposition}{\textbf{Proposition}}

\newtheorem{Example}{\textbf{Example}}

\newcommand{\ie}{\mbox{$i.e.$}}

\newcommand{\baf}{\textit{BAF}\xspace}
\newcommand{\bipolar}{{\ensuremath{\langle \ard, \atts, \supp \rangle}}}
\newcommand{\ard}{\ensuremath{\tt{Arg}}\xspace}
\newcommand{\R}{\ensuremath{\mathrm{R}}}
\newcommand{\atts}{\ensuremath{\tt{R}_{a}}\xspace}
\newcommand{\supp}{\ensuremath{\tt{R}_{s}}\xspace}

\newcommand{\argum}[1]{\ensuremath{\mathcal{#1}}}
\newcommand{\argu}[1]{\ensuremath{\mathcal{#1}}}
\newcommand{\graphbipolar}[1]{\ensuremath{\mathrm{G}_{#1}}\xspace}

\newcommand{\tbaf}{\textit{T\mbox{-}BAF}\xspace}
\newcommand{\av}{\ensuremath{\mathtt{Av}}\xspace}
\newcommand{\timebipolar}{{\ensuremath{\langle \ard, \atts, \supp, \av \rangle}}}

\newcommand{\tprotxt}{\ensuremath{\mathit{t\mbox{-}profile}}\xspace}
\newcommand{\tpro}[1]{\ensuremath{\langle \argu{#1},\mathcal{T}_{\argu{#1}} \rangle}}
\newcommand{\tpron}[2]{\ensuremath{\langle \argu{#1}_{#2},\mathcal{T}_{\argu{#1}_{#2}}\rangle}}
\newcommand{\tproprima}[1]{\ensuremath{\langle \argu{#1},\mathcal{T}'_{\argu{#1}}\rangle}}
\newcommand{\tprobasic}[1]{\ensuremath{\langle \argu{#1},\av(\argu{#1})\rangle}}
\newcommand{\tproins}[2]{\ensuremath{\langle \argu{#1},\{#2\}}\rangle}

\newcommand{\tiempobasic}[1]{\ensuremath{\av(\argu{#1})}}
\newcommand{\tiempo}[1]{ \ensuremath{\mathcal{T}_{ \argu{#1} } } }
\newcommand{\timedefeat}[2]{ \ensuremath{\mathcal{T}^d_{(\argu{#1},\argu{#2})} } }
\newcommand{\timesupport}[2]{ \ensuremath{\mathcal{T}^s_{(\argu{#1},\argu{#2})} } }

\newcommand{\tiempoprima}[1]{ \ensuremath{\mathcal{T}'_{ \argu{#1} } } }
\newcommand{\tiempon}[2]{ \ensuremath{\mathcal{T}_{ \argu{#1}_{#2} } } }
\newcommand{\tprostxt}{\ensuremath{\mathit{t\mbox{-}profiles}}\xspace}

\newcommand{\cutprofiles}[2]{\ensuremath{\prod_{#1}({#2})}}
\newcommand{\tdefence}[2]{\ensuremath{\mathcal{T}_{(\argu{#1} \mid C)}^{\argu{#2}}}}
\newcommand{\tdefenceBis}[3]{\ensuremath{\mathcal{T}_{(\argu{#1} \mid #3)}^{\argu{#2}}}}
\newcommand{\tsupdefence}[2]{\ensuremath{\mathcal{T}_{(\argu{#1} \mid C \mid Sup)}^{\argu{#2}}}}
\newcommand{\tsecdefence}[2]{\ensuremath{\mathcal{T}_{(\argu{#1} \mid C \mid Sec)}^{\argu{#2}}}}
\newcommand{\tdefences}[1]{\ensuremath{\mathcal{T}_{(\argu{#1} \mid C)}}}
\newcommand{\tdefencesBis}[2]{\ensuremath{\mathcal{T}_{(\argu{#1} \mid #2)}}}
\newcommand{\tdefencese}[1]{\ensuremath{\mathcal{T}_{(\argu{#1} \mid C)}}}
\newcommand{\tsuppdefeat}[2]{\ensuremath{\mathcal{T}_{({\argu{#1}}\mbox{-}{\argu{#2}})}^{Sup}}}
\newcommand{\tsecdefeat}[2]{\ensuremath{\mathcal{T}_{({\argu{#1}}\mbox{-}{\argu{#2}})}^{Sec}}}